\documentclass[runningheads]{llncs}

\usepackage{graphicx}
\usepackage{subfigure}
\usepackage{booktabs}
\usepackage{multirow}
\usepackage{adjustbox}
\usepackage{wrapfig}
\usepackage{float}
\usepackage[title]{appendix}
\usepackage{lscape}

\begin{document}
\title{Investigating the Effectiveness of Representations Based on Word-Embeddings in Active Learning for Labelling Text Datasets\thanks{Supported by organization x.}}
%
%
\author{Jinghui Lu\inst{1} \and
Maeve Henchion\inst{2} \and
Brian Mac Namee\inst{1}}

%
%
\institute{Insight Centre for Data Analytics, Univeristy College Dublin, Ireland \and
Teagasc Agriculture and Food Development Authority, Ireland
\email{Jinghui.lu@ucd.connect.ie},
\email{Maeve.henchion@teagasc.ie},
\email{Brian.MacNamee@ucd.ie}}
%

\maketitle         
\begin{abstract}
Manually labelling large collections of text data is a time-consuming and expensive task, but one that is necessary to support machine learning based on text datasets. \emph{Active learning} has been shown to be an effective way to alleviate some of the effort required in utilising large collections of unlabelled data for machine learning tasks without needing to fully label them. The representation mechanism used to represent text documents when performing active learning, however, has a significant influence on how effective the process will be. While simple vector representations such as \emph{bag-of-words} have been shown to be an effective way to represent documents during active learning, the emergence of representation mechanisms based on the \emph{word embeddings} prevalent in neural network research (e.g. \emph{word2vec} and transformer based models like \emph{BERT}) offer a promising, and as yet not fully explored, alternative. This paper describes a large-scale evaluation of the effectiveness of different text representation mechanisms for active learning across 8 datasets from varied domains. This evaluation shows that using representations based on modern word embeddings, especially BERT, which have not yet been widely used in active learning, achieves a significant improvement over more commonly used vector representations like bag-of-words.

\keywords{active learning  \and text classification \and word embeddings \and BERT \and FastText}
\end{abstract}
\section{Introduction}\label{sec:intro}

 \emph{Active learning} (AL) \cite{settles2009active} is a semi-supervised machine learning technique that minimises the amount of labelled data required to build accurate prediction models. In active learning only the most informative instances from an unlabelled dataset are selected to be labelled by an oracle (i.e. a human annotator) to expedite the learning procedure. This property makes active learning attractive in scenarios where unlabelled data may be abundant but labelled data is expensive to obtain such as image classification \cite{tong2001support,zhang2002active}, speech recognition \cite{tur2005combining}, and text classification \cite{hoi2006large,Liere1997active,zhang2017active,singh2018improving}---which is the focus of this work. 

One crucial component of active learning for text classification is the mechanism used to represent documents in the tabular structure required by most machine learning algorithms. 
Vectorized representations, such as \emph{bag-of-words} (BOW) is the most common representations used in active learning \cite{singh2018improving,hu2010egal,hu2008sweetening,wallace2010active,siddhant2018deep,miwa2014reducing}. 
Considerable recent work, however, has shown that representations of natural language based on learned \emph{word embeddings} can be useful for a wide range of natural language processing (NLP) tasks including text classification \cite{mikolov2013efficient,pennington2014glove,bojanowski2017enriching,howard2018universal,radford2018improving,devlin2018bert,peters2018deep}. Standard approaches to learning word embeddings like \emph{word2vec} \cite{mikolov2013efficient}, \emph{Glove} \cite{pennington2014glove}, \emph{FastText} \cite{bojanowski2017enriching,joulin2016bag}, or contextualized approaches such as \emph{Cove} \cite{mccann2017learned} and \emph{ElMo} \cite{peters2018deep} convert words to fixed-length dense vectors that capture semantic and syntactic features, and allow more complex structures (like sentences, paragraphs and documents) to be encoded as aggregates of these vectors. There are also document-level approaches such as \emph{ULM-Fit} \cite{howard2018universal}, \emph{OpenAI GPT} \cite{radford2018improving}, and \emph{BERT} \cite{devlin2018bert}, that are followed by task-specific fine-tuning to significantly increase the performance of NLP tasks, and have been shown to be useful for learning common language features. Among these approaches, BERT (\emph{bidirectional encoder representations from transformers}) has achieved state-of-the-art results across many NLP tasks \cite{devlin2018bert}. Notably, in these approaches language models used for generating embedding-based representations are created using massive unlabelled corpora (e.g. Wikipedia \cite{mikolov2013efficient,devlin2018bert}). Even though word embeddings have been widely applied in text classification, there is little work devoted to leveraging them in active learning for text classification \cite{zhang2017active,zhao2017deep,siddhant2018deep}, and a large-scale benchmark comparison of their usefulness for active learning does not exist in the literature.  
 

This paper describes a large-scale evaluation experiment to explore the effectiveness of word embeddings for active learning in a text classification context. This evaluation, based on 8 datasets from different domains such as product reviews, news articles, blog posts etc., shows that representations based on word-embeddings---and especially representations based on BERT---consistently outperform the more commonly used simple vector representations. This demonstrates the effectiveness of embedding-based representations for active learning, and also illustrates that some of the promise of deep learning \cite{lecun2015deep} can be brought to active learning, while avoiding the considerable practical challenges of placing a deep neural network at the heart of the active learning process \cite{zhang2017active,zhao2017deep,siddhant2018deep,zhang2019neural}. 

The remainder of the paper is organized as follows: Section \ref{sec:rel_work} presents the pool-based active learning framework as well as the text representation techniques and related work; Section \ref{sec:experiments} describes the design of the experiment performed; Section \ref{sec:results} discusses the results of this experiment; and, finally, Section \ref{sec:conclusions} draws conclusions and suggests directions for future work.

\section{Related Work}\label{sec:rel_work}

In \emph{pool-based} active learning, a small set of labelled instances is used to seed an initial labelled dataset, $\mathcal{L}$. Then, according to a particular selection strategy, a batch of data to be presented to an oracle for labelling is chosen from the unlabelled data pool, $\mathcal{U}$. After labelling, these newly labelled instances will be removed from $\mathcal{U}$ and appended to $\mathcal{L}$. This process repeats until a predefined stopping criterion has been met  (e.g. a label budget has been exhausted). 
 
The resulting labelled instances will be used for training a predictive model if the goal is to induce a good classification model; if the goal is to label all instances, the induced classification model will be applied to the remaining unlabelled instances in $\mathcal{U}$ to predict their classes, which saves manual labours as compared to labelling the whole dataset by hand. We are interested in the latter scenario in this paper.

Selection strategy, which is a technique used for picking unlabelled data to be presented to the oracle for labelling, plays a vital role in active learning. There are many selection strategies studied in the literature. A family of approaches, such as uncertainty sampling, query-by-committee (QBC), density-weighted methods \cite{settles2009active}, utilise models trained with the currently labelled instances, $\mathcal{L}$, to infer the ``informativeness'' of unlabelled instances from $\mathcal{U}$, among which the most informative are selected to be labelled by the oracle. We refer to these approaches as \emph{model-based} selection strategies. On the other hand, there are methods entirely rely on the features of instances in $\mathcal{L}$ and $\mathcal{U}$ to compute the ``informativenes'' of each candidate such as exploration guided active learning (EGAL) \cite{hu2010egal}, which is referred to as \emph{model-free} selection strategy \cite{hu2010egal}. In this paper, we adopt several commonly used selection strategies, that is, \emph{random sampling} (sample i.i.d from $\mathcal{U}$), \emph{uncertainty sampling} \cite{lewis1994sequential}, \emph{query-by-committee} \cite{seung1992query}, \emph{information-density (ID)} \cite{settles2008analysis}, \emph{EGAL} \cite{hu2010egal} to alleviate the influence caused by different selection strategies.

\subsection{Text Representations}

Choosing a representation scheme for documents to be used in an active learning scenario is an important step. Approaches to doing this range from simple frequency based vector representations, like bag-of-words, to more sophisticated approaches based on word embeddings. This section presents the most text representation schemes used in the experiments described in this paper.

\noindent \textbf{Bag-of-words} (BOW) is the most basic representation scheme for documents, and has been widely used in many active learning applications \cite{singh2018improving,hu2010egal,hu2008sweetening,wallace2010active,siddhant2018deep,miwa2014reducing}. Each column of the BOW vector is the term-frequency (TF) of a distinct word appearing in the document and 0 if the term is absent. The frequency of terms is often weighted by inverse document frequency to penalise terms commonly used in most documents. This is known as TF-IDF \cite{jones2004statistical}. In this paper, we examine both TF-IDF, and TF which is normalized by the total word count of a document.

\noindent \textbf{Latent Dirichlet Allocation} (LDA) \cite{blei2003latent} is a topic modelling techniques technique designed to infer the distribution of membership to a set of topics across a set of documents. The model generates a term-topic matrix and a document-topic matrix. Specifically, each row of the document-topic matrix is a topic-based representation of a document where the $i$th column determines the degree of association between the $i$th topic and the document. This type of topic representation of documents has been used in active learning for labelling inclusive/exclusive studies in systematic literature review \cite{singh2018improving,hashimoto2016topic,mo2015supporting}.

\noindent \textbf{FastText} \cite{bojanowski2017enriching} is a neural language model trained with large online unlabelled corpora. Compared to word2vec \cite{mikolov2013efficient} and Glove \cite{pennington2014glove}, FastText enriches the training of word embeddings with subword information which improves the ability to obtain word embeddings of out-of-bag words. In practice, we average the vectors of words appeared in the document as the document representation.

\noindent \textbf{BERT} \cite{devlin2018bert} has achieved amazing results in many NLP tasks. This model, using multi-head attention mechanism based on the Multi-layer Bidirectional Transformer model \cite{vaswani2017attention}, is trained with the plain text through masked word prediction and next sentence prediction tasks to learn contextualized word embeddings. Contextualized word embedding implies a word can have different embeddings according to its context which alleviates the problems caused by polysemy etc.

\subsection{Using Word Embedding in Active Learning}

Although applying word embeddings in text classification has attracted considerable attention in the literature \cite{mikolov2013efficient,bojanowski2017enriching,howard2018universal,devlin2018bert}, the use of word embeddings in active learning is still a largely unexplored research area. Zhang et al. \cite{zhang2017active} combined word2vec with convolutional neural networks (CNN) and active learning to build classifiers for sentence-based and document-based datasets. 
Similarly, Zhao et al. \cite{zhao2017deep} proposed leveraging recurrent neural networks and gated recurrent units with word2vec to predict the classes of short-text. Very recently, Zhang Ye \cite{zhang2019neural} proposed a selection strategy that combines fine-tuned BERT with CNN, but they only compare the performance of different selection strategies while BERT applied, rather than the impacts of different text representation techniques used in active learning. 
Siddhant and Lipton \cite{siddhant2018deep} compare the performance of Glove-embedding-based active learning frameworks, which are composed of different classifiers such as bi-LSTM model and CNN, across many NLP tasks. They find that Glove embeddings selected by Bayesian Active Learning by Disagreement plus Monte Carlo Dropout or Bayes-by-Backprop Dropout usually outperforms the shallow baseline. However, Siddhant and Lipton take Linear SVM combined with BOW representation rather than Glove embeddings as a shallow baseline which makes the conclusion limited.  
Additionally, these four studies focus on comparing the impact of selection strategies when used with deep neural networks, instead of that of text representations.

Another challenge faced by approaches that use word embeddings in combination with deep neural networks is the computational cost in active learning. Highly computationally complex prediction models such as neural networks are too expensive to be used in active learning due to the frequent demand of reconstructing classifiers. Therefore, some studies combine word embedding with low complexity machine learning algorithms to provide tractable approaches. 
Hashimoto et al. \cite{hashimoto2016topic} propose a method, paragraph vector-based topic detection (PV-TD), that combines doc2vec \cite{le2014distributed} (an extension of word2vec) with k-means clustering to perform simple topic modelling. For the active learning process documents, which are represented by their distance to the cluster centres that result from the application of k-means, are fed into the Support Vector Machine (SVM) model. In their experiments PV-TD is shown to perform well compared to representations based on an LDA, and word2vec. 
Interestingly, Singh et al. \cite{singh2018improving} extend the experiments in \cite{hashimoto2016topic} with more datasets in the health domain, demonstrating that directly using doc2vec or BOW, rather than PV-TD, can achieve better results which is contrary to that obtained by Hashimoto.
Despite the promising results, these studies of active learning using word embeddings explore only a limited number of selection strategies (i.e. certainty sampling and certainty information gain sampling) and focus only on highly imbalanced datasets from specialist medical domains. 

This research fills the gap by comparing the performance of embedding-based active learning with that of classical active learning framework, using a broader range of selection strategies and datasets of various domains to fairly demonstrate the effectiveness of each representation. As far as we know, this is the first attempt to evaluate the performance of BERT as a representation compared to other vector-based representations in the context of expediting text labelling tasks via active learning.

\section{Experimental Design}\label{sec:experiments}

This section describes the design of an experiment performed to evaluate the effectiveness of different text representation mechanisms in active learning. To mitigate the influence of different selection strategies on the performance of the active learning process we also include a number of different selection strategies in the experiment. These section describes the experimental framework, the configuration of the models used, the performance measures used to judge the effectiveness of different approaches, and the datasets used in the experiments. 

\subsection{Active Learning Framework}

We apply pool-based active learning using different text representation techniques and selection strategies over several well-balanced fully labelled datasets. All datasets are from binary classification problems. The use of fully labelled datasets allows us to simulate data labelling by a human oracle, and is common in active learning research \cite{zhang2017active,zhao2017deep,singh2018improving,hu2010egal,hashimoto2016topic}. At the outset, we provide all learners with the same 10 instances (i.e. 5 positive instances and 5 negative instances) sampled i.i.d. at random from a dataset to seed the active learning process. Subsequently, 10 unlabelled instances, whose ground truth labels will be revealed to each learner, are selected according to a certain selection strategy. These examples are removed from $\mathcal{U}$ to $\mathcal{L}$ and the classifiers are retrained. We assume it is unrealistic to collect more than 1,000 labels from an oracle, and so we stop the procedure when an annotation budget of 1,000 labels is used up. As the batch size for selection is 10, this means that an experiment is composed of 100 \emph{rounds} of the active learning process which uses up the label budget of 1,000 labels.  Each experiment is repeated 10 times using different random seeds. The performance measures reported are averaged across these repetitions.

\subsection{Model Configuration}

We evaluate the performance of active learning using Linear-SVM models\footnote{https://scikit-learn.org/stable/modules/generated/sklearn.svm.SVC.html} which have been shown empirically to perform well with high dimensional data \cite{hsu2003practical}. We tune the hyper-parameters of the SVM models every 10 iterations (i.e. 100 labels requested). We preprocess text data by converting to lowercase, removing stop words, and removing rare terms (for the whole dataset, word count less than 10 or document frequency less than 5).\footnote{We found the preprocessing improves the performance of BOW but has a negligible effect to word embeddings, hence we skip preprocessing for word embeddings.} We set the number of topics to be used by LDA\footnote{https://radimrehurek.com/gensim/models/ldamodel.html} to 300 following \cite{singh2018improving}. For FastText, we adopt two versions of FastText: 1) we use the pre-trained subword FastText (FT) model trained with Wikipedia (300 dimensions).\footnote{https://fasttext.cc/docs/en/english-vectors.html} 2) continually training original FastText model with local corpus (without label information) which is referred to as FastText\_trained (FT\_T). For BERT, we use bert-large-uncased model (1,024 dimensions).\footnote{https://github.com/huggingface/pytorch-transformers} Though the original paper \cite{devlin2018bert} suggests using the vector of ``[CLS]'' token added in the head of a document as a document-level representation, in practice, researchers find that averaging the word embeddings of the document is an equivalent, sometimes, greater option \footnote{https://github.com/hanxiao/bert-as-service}. Since BERT is configured to take as input a maximum of 512 tokens, we divided the long sequence with $L$ length into $k = L/511$ fractions, which is then fed to BERT to infer the representation of each fraction (each fraction has ``[CLS]'' token in front of 511 tokens, namely, 512 tokens in total). The vector of each fraction is the average embeddings of words in that fraction and the representation of the whole text sequence is the mean of all $k$ fraction vectors. It should be noted that we do not use any label information for fine-tuning any model to ensure fair comparisons. A summary of the dimensionality of each representation is given in Table \ref{tab:datasets}.

In uncertainty sampling, the most uncertain examples are equivalent to those closest to the class separating hyper-plane in the context of an SVM \cite{tong2000support}. In the information density selection strategy, we use entropy to measure the ``informativeness'' and all parameters are set following \cite{settles2008analysis}. In QBC, we choose Linear-SVM models trained using bagging as committee members following \cite{mamitsuka1998query}. Since there is no general agreement in the literature on the appropriate committee size for QBC \cite{settles2009active}, we adopt committee size 5 after some preliminary experiments. In EGAL, all parameters are set following the recommendations given in \cite{hu2010egal}, which are shown to perform well for text classification tasks. 

\subsection{Performance Measures}

As we are interested in the ability of an active learning process to fully label a dataset we use the \emph{accuracy+} performance measure, which has been previously used by \cite{hu2010egal}. This measures the performance of the full active learning system including human annotators. It can be expressed as:

\begin{equation}
accuracy+ = \frac{TP^{H}+TN^{H}+TP^{M}+TN^{M}}{N}
\label{eq:acc}
\end{equation}

\noindent where $N$ is the total number of instances in a dataset  and superscripts $H$ and $M$ express human annotator and machine generated labels respectively. $TP$ and $TN$ denote the number of \emph{true positives} and \emph{true negatives} respectively. Intuitively, this metric computes the fraction of correctly labelled instances which are predicted by the oracles as well as a trained classifier. We presume that a human annotator never makes mistakes. We also report the \emph{area under the learning curve} (AULC) score for each accuracy curve which is computed using the trapezoidal rule and normalized by the maximum possible area, to bound the value between 0 to 1.

\subsection{Datasets}

We evaluate the performance of active learning systems using 8 fully-labelled datasets. Four of these datasets are based on long text sequences: \emph{Movie Review (MR)} \cite{pang2004sentimental},\footnote{\textbf{MR and MRS are available at:} http://www.cs.cornell.edu/people/pabo/movie-review-data/} \emph{Multi-domain Customer Review (MDCR)} \cite{blitzer2007biographies},\footnote{https://www.cs.jhu.edu/\textasciitilde mdredze/datasets/sentiment/index2.html} \emph{Blog Author Gender (BAG)} \cite{mukherjee2010improving}\footnote{\textbf{BAG and ACR are available at:} https://www.cs.uic.edu/\textasciitilde liub/FBS/sentiment-analysis.html} and \emph{Guardian2013 (G2013)} \cite{belford2018stability}. While four are based on sentences: \emph{Additional Customer Review (ACR)} \cite{ding2008holistic}, \emph{Movie Review Subjectivity (MRS)} \cite{pang2004sentimental}, \emph{Ag news (AGN)}\footnote{\textbf{AGN and DBP are available at :} https://skymind.ai/wiki/open-datasets} and \emph{DBP(Dbpedia)} \cite{zhang2015character}. Table \ref{tab:datasets} provides summary statistics describing each dataset.

\begin{table}[!htbp] 
\caption{Statistics of 8 datasets. Left column set denotes the number of positives and negatives in each dataset, right column set denotes the vector length of different representations \emph{wrt.} each dataset. FT and FT\_T denote FastText and FastText\_trained.}
\label{tab:datasets}
\small
    \centering
\begin{tabular}{l|rr|r r r r r r } \hline \hline
      \multicolumn{1}{c|}{} &
      \multicolumn{2}{c|}{\# of Instances}& \multicolumn{6}{c}{Representation Dimensionality} \\ 
     {\fontseries{b}\selectfont Dataset} & {\fontseries{b}\selectfont positives}      & {\fontseries{b}\selectfont negatives} & { \fontseries{b}\selectfont  TF} & {\fontseries{b}\selectfont TFIDF} & {\fontseries{b}\selectfont LDA} & 
     {     \fontseries{b}\selectfont   FT} & {\fontseries{b}\selectfont  FT\_T} & {\fontseries{b}\selectfont BERT}  \\   \hline
MR         & 1,000          & 1,000          & 6,181 & 6,181& 300 & 300 & 300  & 1,024  \\
MDCR      &     4,000          & 3,566          & 4,165 & 4,165& 300 & 300 & 300  & 1,024  \\
BAG        &    1,675          & 1,552          & 4,936 & 4,936& 300 & 300 & 300  & 1,024  \\
G2013     &       843          & 1,292          & 5,345 & 5,345& 300 & 300 & 300  & 1,024  \\ \hline

ACR       &    1,335          & 736         &403 & 403& 300 & 300 & 300  & 1,024  \\

MRS     &      5,000          & 5,000         & 1,868 & 1,868& 300 & 300 & 300  & 1,024 \\ 
AGN     &      1,000          & 1,000         & 723 & 723& 300 & 300 & 300  & 1,024  \\
DBP     &      1,000          & 1,000         & 552 & 552& 300 & 300 & 300  & 1,024  \\
\hline\hline

\end{tabular}
\vspace{-5mm}
\end{table}

\begin{figure*}[!htbp]
  \centering
  \subfigure[Random]{\includegraphics[width=0.32\linewidth]{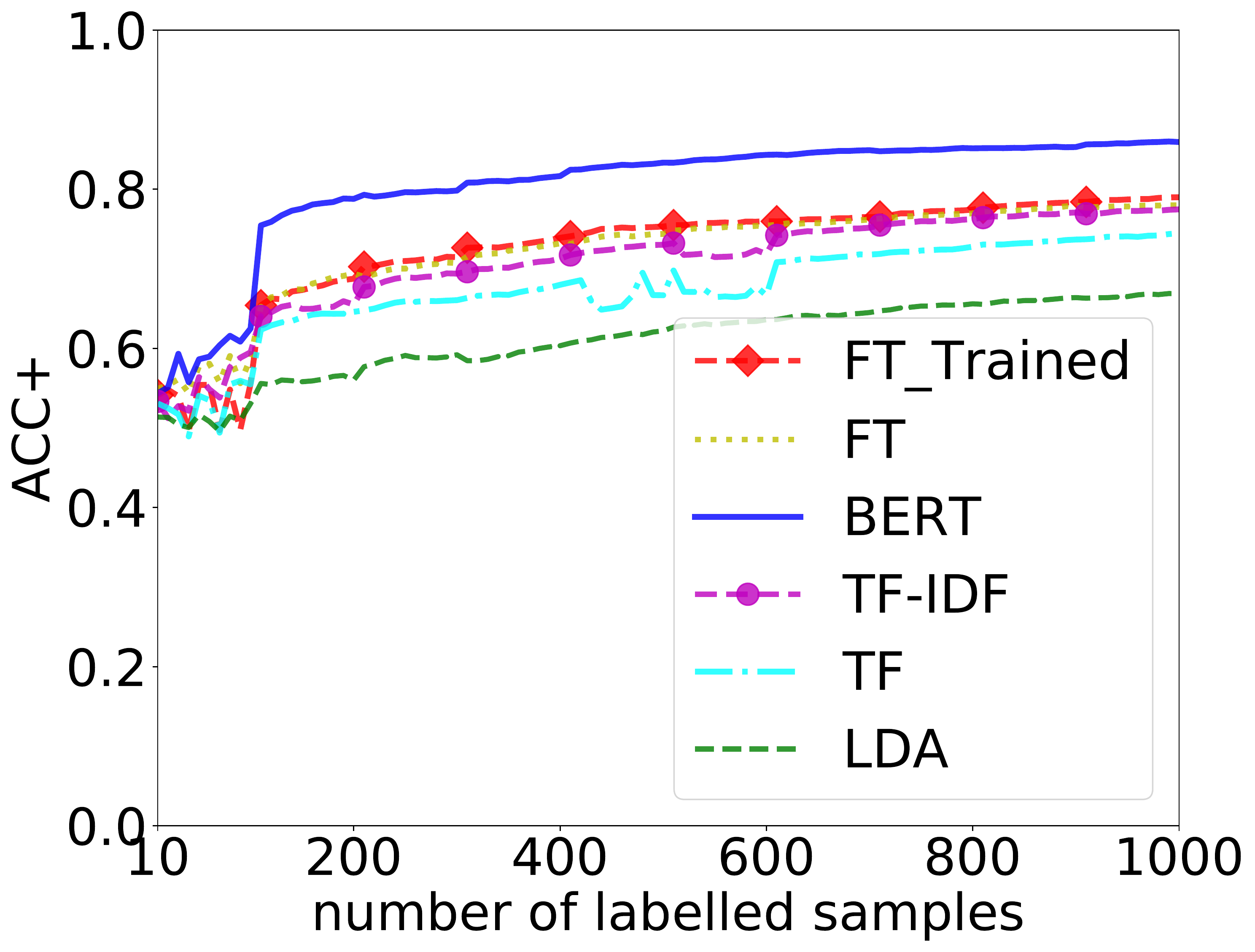}}
  \subfigure[Uncertainty]{\includegraphics[width=0.32\linewidth]{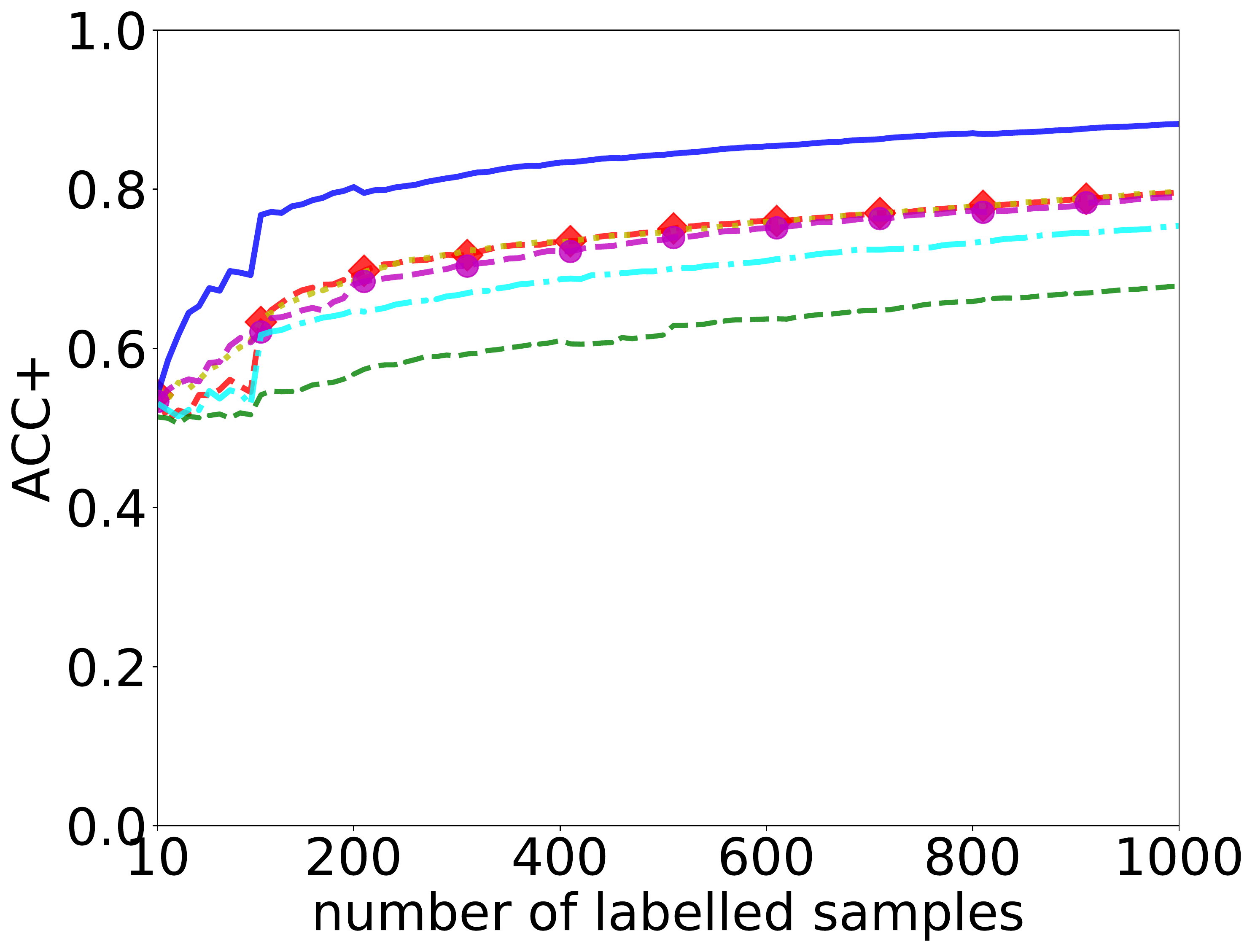}}
\subfigure[Information Density]{\includegraphics[width=0.32\linewidth]{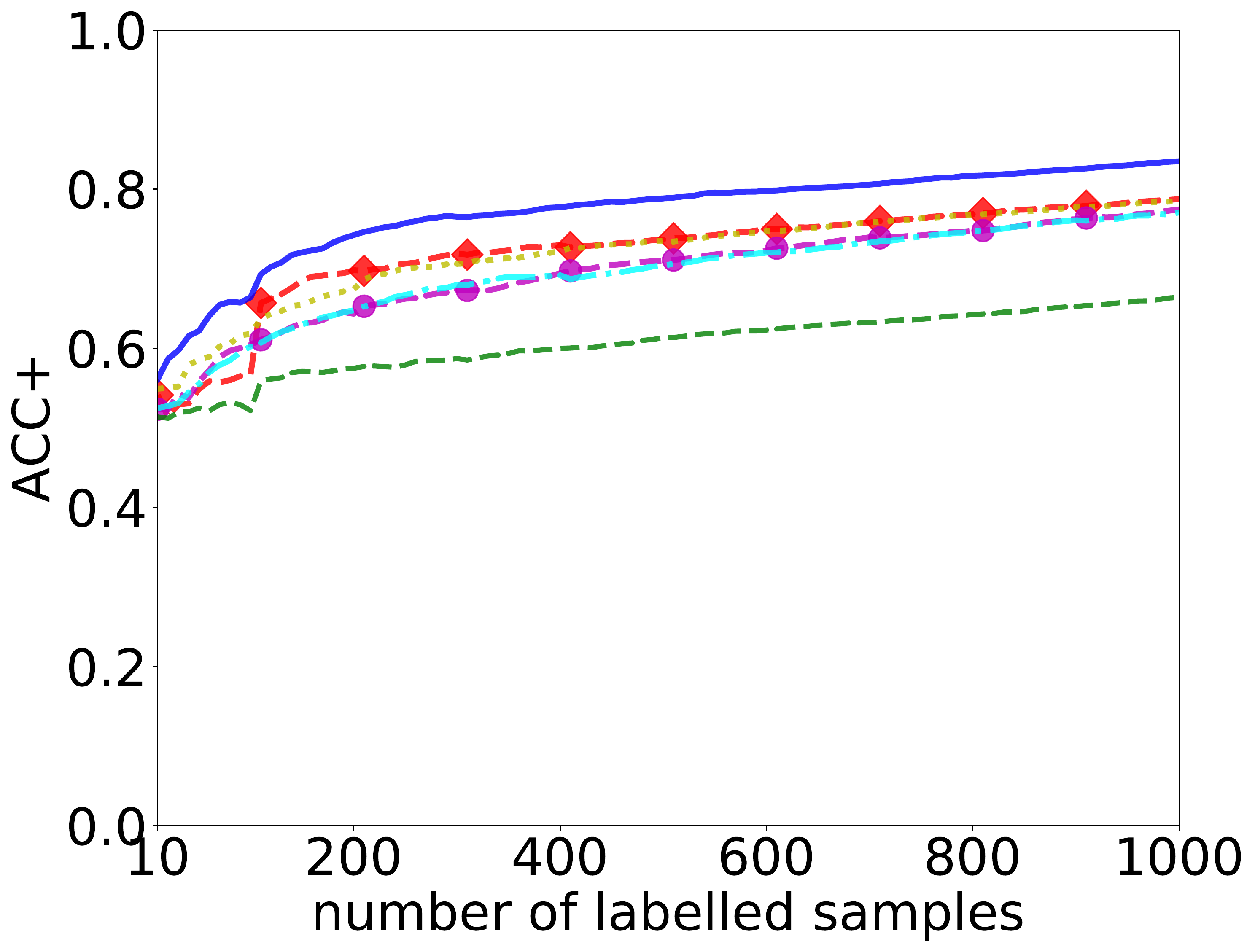}}
\medskip
  \centering
  \subfigure[QBC]{\includegraphics[width=0.32\linewidth]{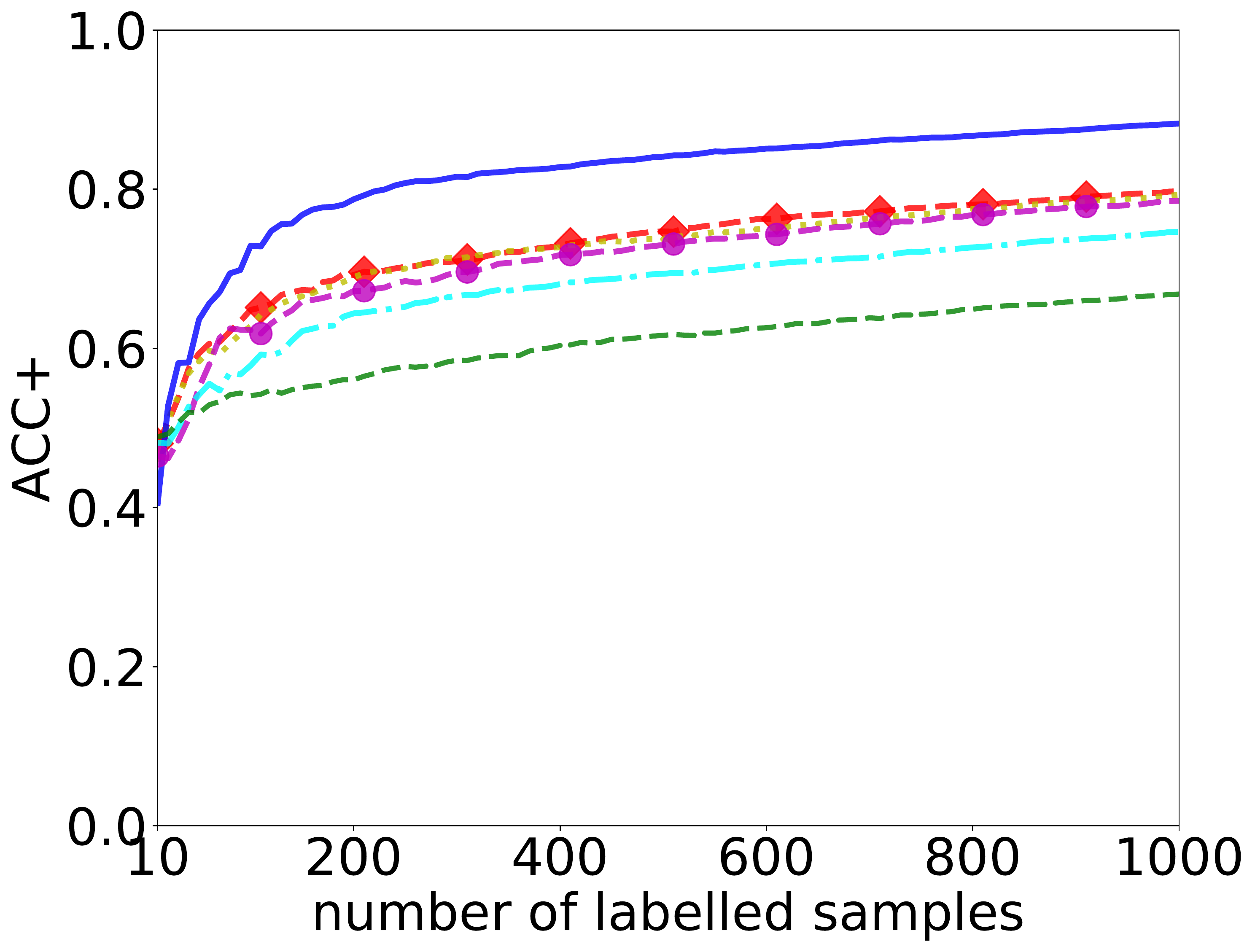}}
  \subfigure[EGAL \label{fig:MDCR_EGAL}]{\includegraphics[width=0.32\linewidth]{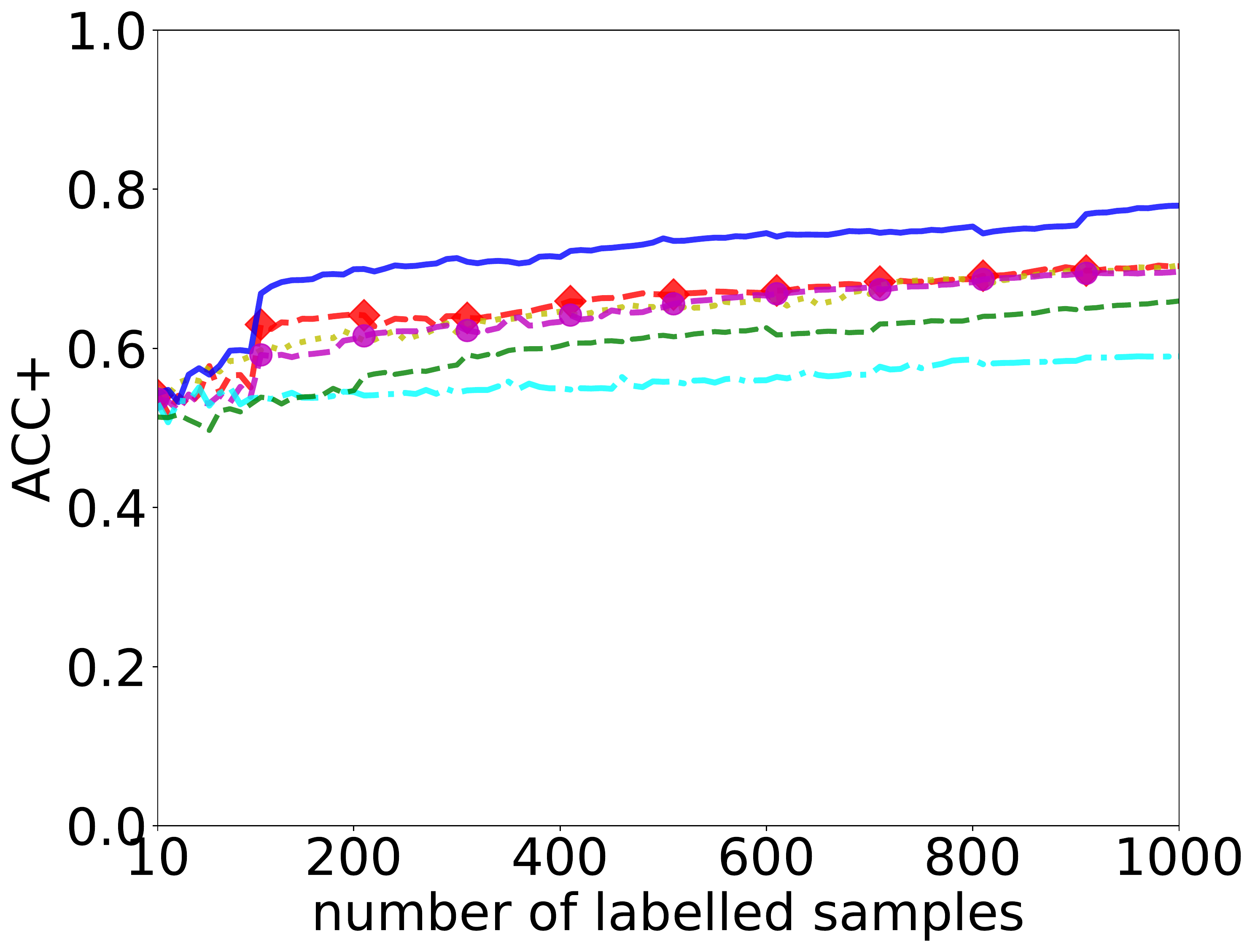}}
  \caption{Results over \emph{Multi-Domain Customer Review (MDCR)} dataset. X-axis represents the number of documents that have been manually annotated and Y-axis denotes accuracy+. Each curve starts with 10 along X-axis.}
  \label{fig:MDCR}
  
\end{figure*}

\begin{figure*}[!htbp]
  \centering
  \subfigure[Random]{\includegraphics[width=0.328\linewidth]{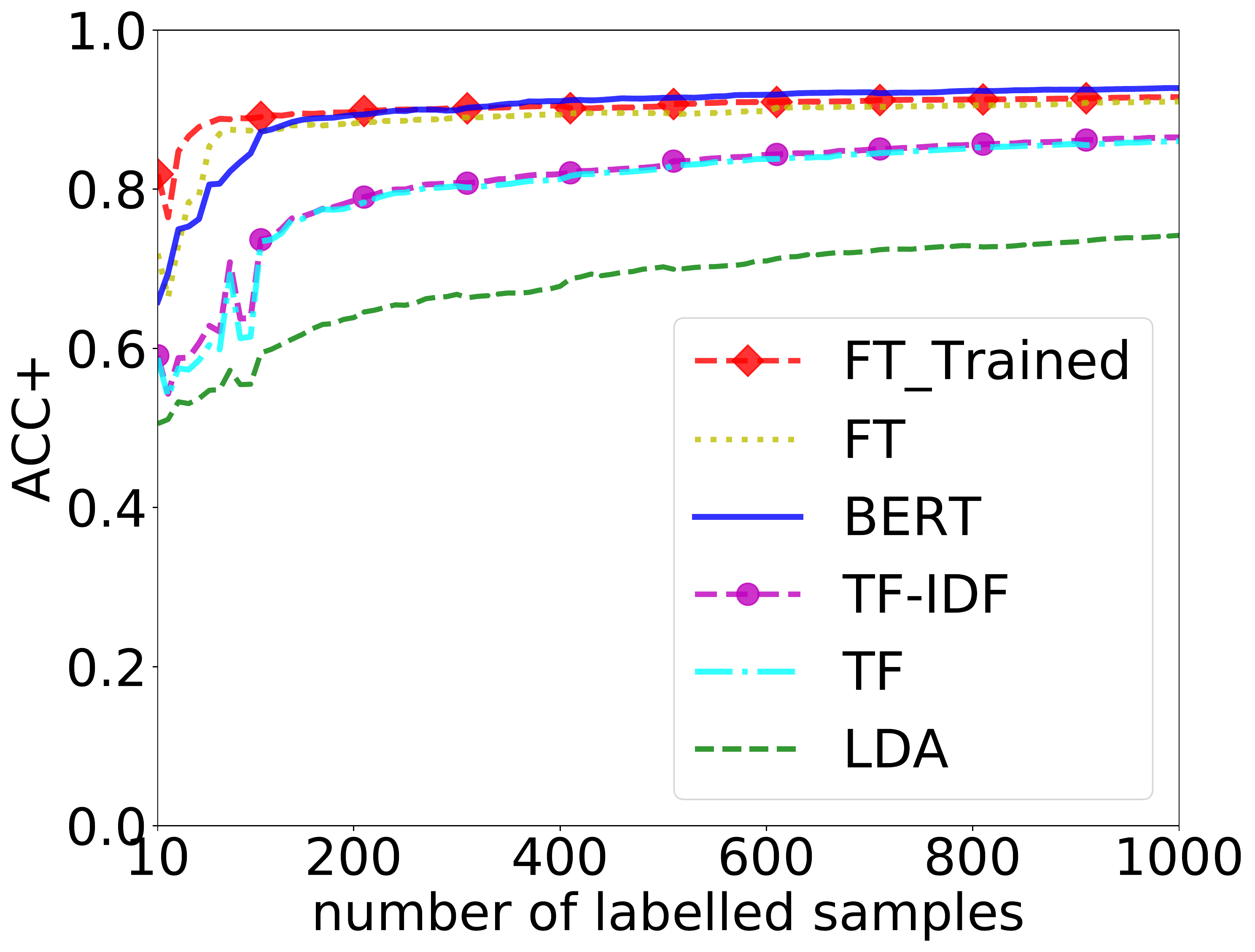}}
  \subfigure[Uncertainty]{\includegraphics[width=0.328\linewidth]{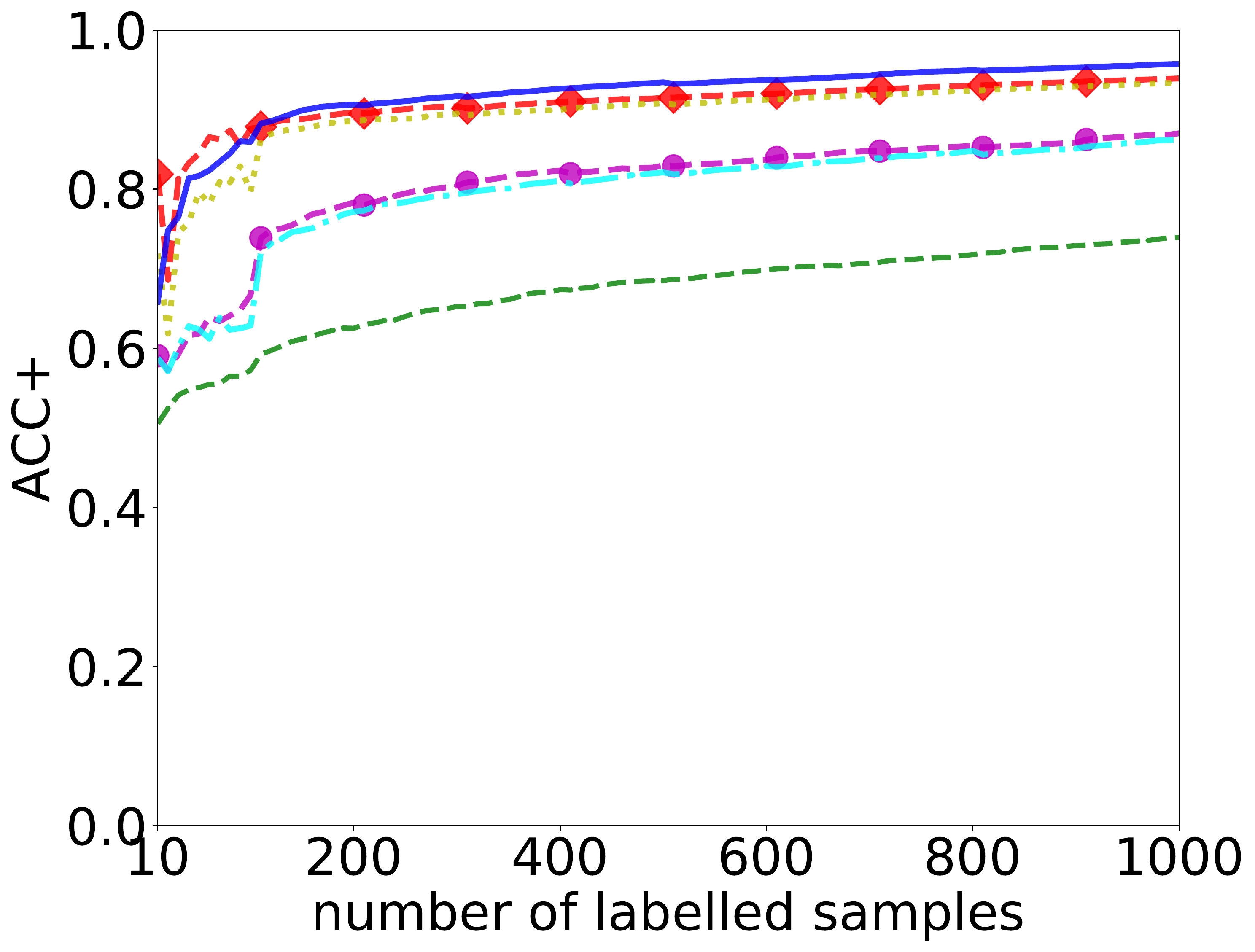}}
\subfigure[Information Density]{\includegraphics[width=0.328\linewidth]{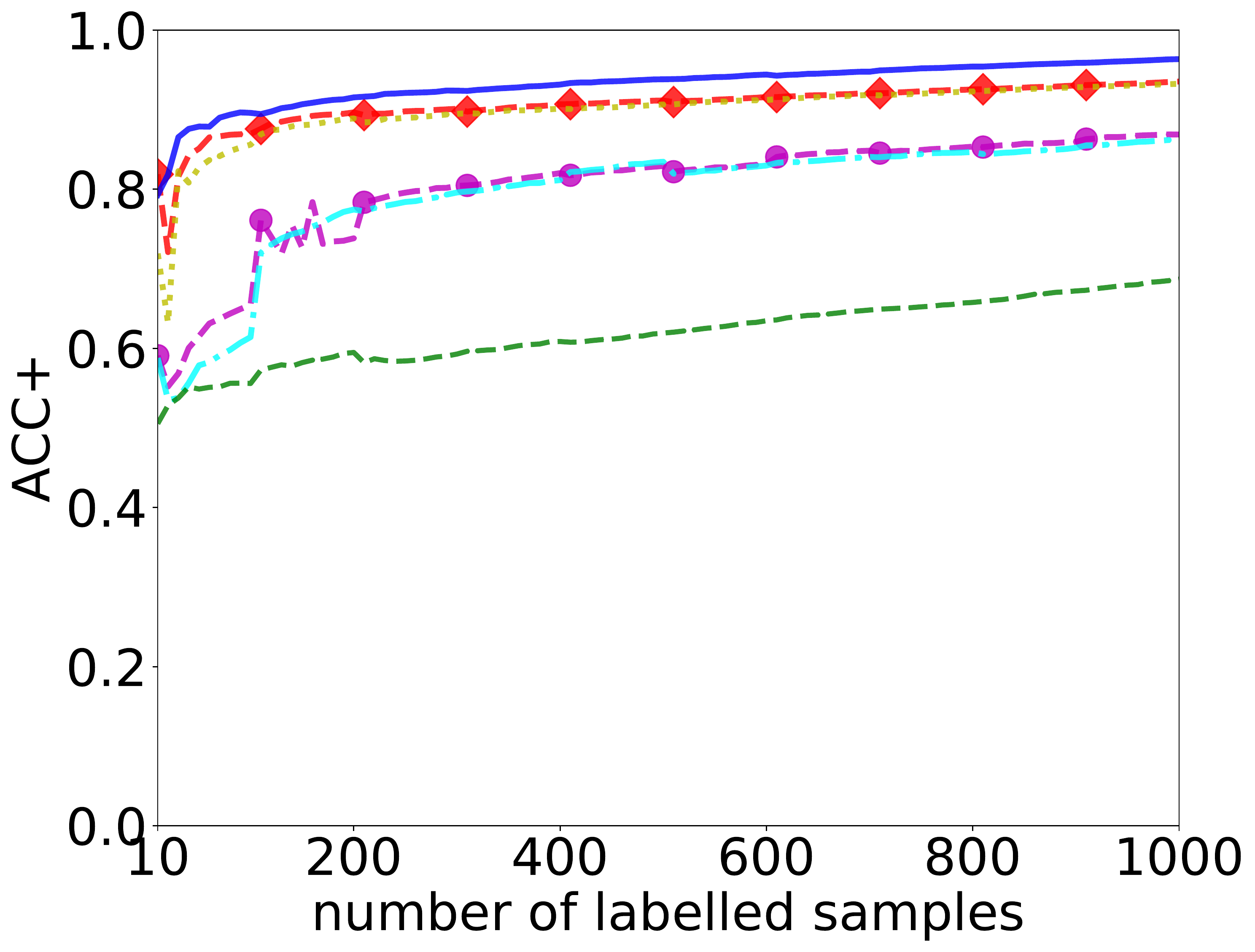}}
\medskip
  \centering
  \subfigure[QBC]{\includegraphics[width=0.328\linewidth]{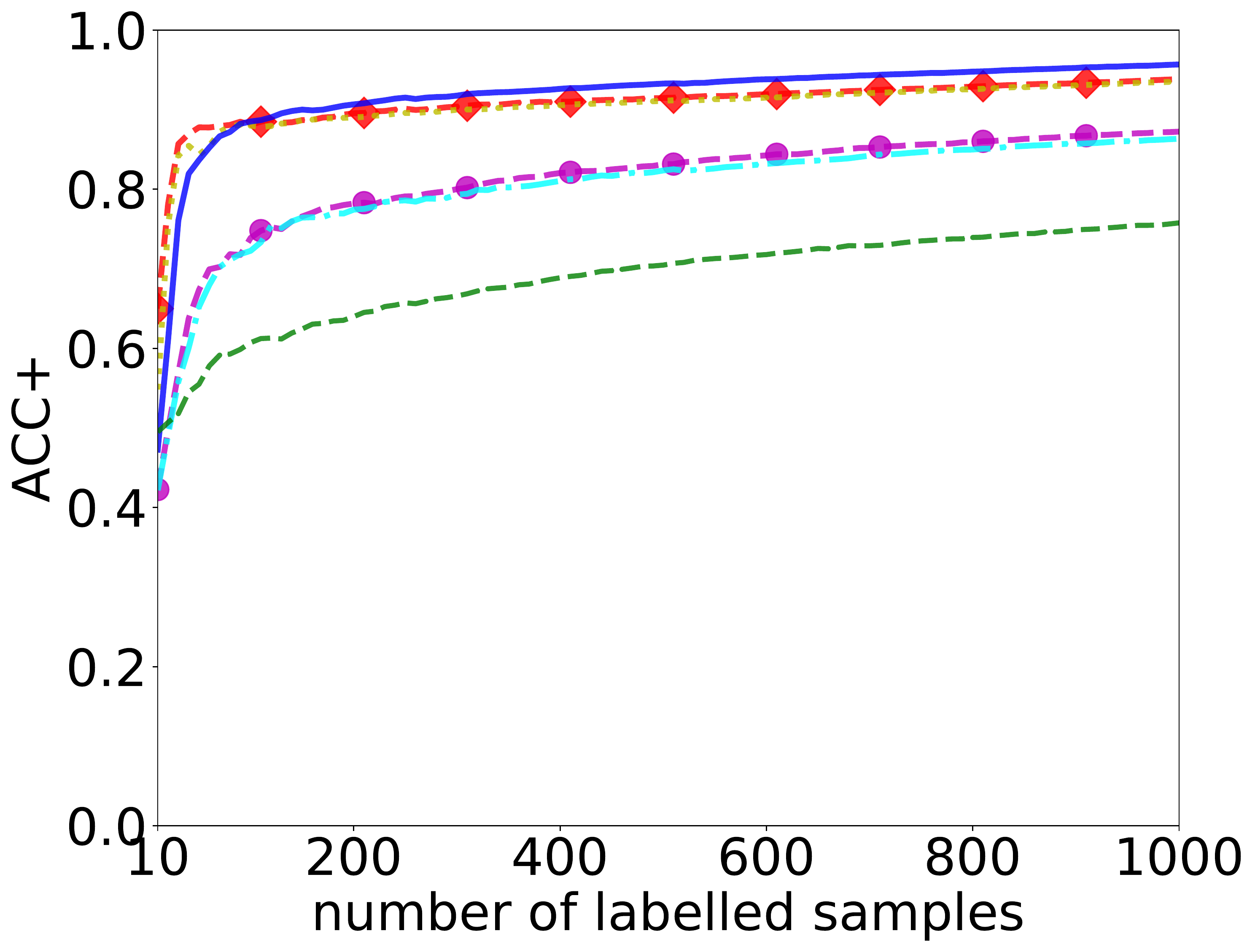}}
  \subfigure[EGAL \label{fig:MRS_EGAL}]{\includegraphics[width=0.328\linewidth]{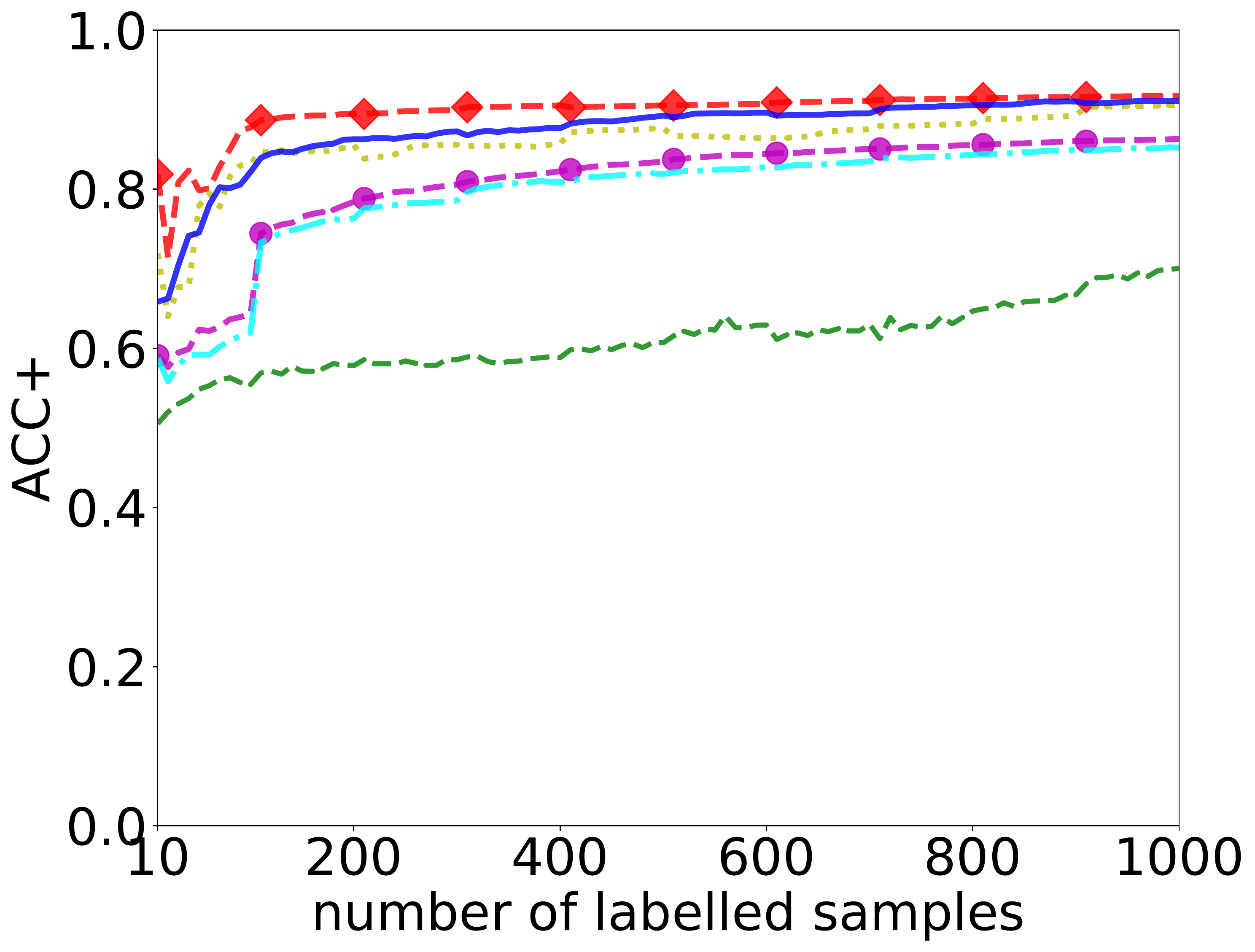}}
  \caption{Results over \emph{Movie Review Subjectivity (MRS)} dataset. X-axis represents the number of documents that have been manually annotated and Y-axis denotes accuracy+. Each curve starts with 10 along X-axis.}
  \label{fig:MRS}

\end{figure*}

\begin{landscape}
\renewcommand{\arraystretch}{1.1}
\begin{table}[!htbp]
\vspace*{-\baselineskip}
\centering
\caption{The summary of AULC scores, which are computed by trapezoidal rule and normalized by the maximum possible area, on each dataset regarding the different combinations of text representations and selection strategies. The best performance of each dataset is highlighted and the last column denotes the average ranking of each method where the smaller number suggests a higher rank.}
\label{tab:AUC_summary}

\resizebox{1.5\textwidth}{!}{%

\begin{tabular}{l|l|l|l|l|l|l|l|l|l|r}
\hline\hline
\textbf{Rep} & \textbf{Strategy} & \textbf{MR} & \textbf{MDCR} & \textbf{BAG} & \textbf{G2013} & \textbf{ACR} & \textbf{MRS} & \textbf{AGN} & \textbf{DBP} & \textbf{Rank} \\\hline
\multirow{5}{*}{BERT} & \textit{random} & 0.857$\pm$0.022(7.0) & 0.805$\pm$0.009(3.0) & 0.717$\pm$0.008(3.0) & 0.946$\pm$0.002(19.0) & 0.789$\pm$0.003(5.0) & 0.899$\pm$0.011(9.0) & 0.962$\pm$0.003(11.0) & 0.976$\pm$0.006(14.0) & 8.88 \\
 & \textit{uncertainty} & \textbf{0.897$\pm$0.024(1.0)} & \textbf{0.823$\pm$0.009(1.0)} & 0.728$\pm$0.009(2.0) & 0.976$\pm$0.001(4.5) & 0.823$\pm$0.007(2.0) & 0.920$\pm$0.009(2.0) & 0.985$\pm$0.002(3.0) & 0.989$\pm$0.003(7.0) & \textbf{2.81} \\
 & \textit{ID} & 0.857$\pm$0.008(8.0) & 0.772$\pm$0.004(4.0) & \textbf{0.735$\pm$0.005(1.0)} & 0.975$\pm$0.002(7.0) & 0.822$\pm$0.009(3.0) & \textbf{0.932$\pm$0.001(1.0)} & 0.985$\pm$0.001(4.0) & 0.988$\pm$0.004(8.0) & 4.50 \\
 & \textit{EGAL} & 0.853$\pm$0.025(11.0) & 0.716$\pm$0.030(14.0) & 0.665$\pm$0.011(23.0) & 0.941$\pm$0.003(24.0) & 0.769$\pm$0.009(13.0) & 0.875$\pm$0.006(14.0) & 0.957$\pm$0.007(14.0) & 0.973$\pm$0.006(15.0) & 16.00 \\
 & \textit{QBC} & 0.892$\pm$0.026(2.0) & 0.818$\pm$0.010(2.0) & 0.714$\pm$0.010(4.0) & 0.971$\pm$0.002(11.5) & \textbf{0.830$\pm$0.005(1.0)} & 0.919$\pm$0.009(3.0) & 0.982$\pm$0.001(6.0) & 0.988$\pm$0.000(9.0) & 4.81 \\\hline
\multirow{5}{*}{FT} & \textit{random} & 0.821$\pm$0.006(20.0) & 0.724$\pm$0.007(9.5) & 0.705$\pm$0.006(8.0) & 0.950$\pm$0.002(16.0) & 0.740$\pm$0.011(24.5) & 0.888$\pm$0.003(13.0) & 0.953$\pm$0.002(15.0) & 0.979$\pm$0.007(13.0) & 14.88 \\
 & \textit{uncertainty} & 0.853$\pm$0.008(12.0) & 0.728$\pm$0.014(6.0) & 0.701$\pm$0.018(10.0) & 0.980$\pm$0.003(2.0) & 0.776$\pm$0.018(8.0) & 0.894$\pm$0.011(12.0) & 0.979$\pm$0.005(9.0) & 0.992$\pm$0.004(5.0) & 8.00 \\
 & \textit{ID} & 0.852$\pm$0.008(14.0) & 0.720$\pm$0.008(12.0) & 0.697$\pm$0.015(12.0) & \textbf{0.981$\pm$0.002(1.0)} & 0.776$\pm$0.012(9.0) & 0.898$\pm$0.006(10.0) & 0.981$\pm$0.004(7.0) & 0.990$\pm$0.006(6.0) & 8.88 \\
 & \textit{EGAL} & 0.814$\pm$0.005(22.0) & 0.647$\pm$0.041(23.0) & 0.656$\pm$0.018(26.0) & 0.947$\pm$0.003(17.0) & 0.737$\pm$0.009(26.0) & 0.859$\pm$0.031(15.0) & 0.961$\pm$0.004(12.0) & 0.980$\pm$0.007(12.0) & 19.12 \\
 & \textit{QBC} & 0.857$\pm$0.005(9.0) & 0.724$\pm$0.007(9.5) & 0.706$\pm$0.013(6.0) & 0.976$\pm$0.001(4.5) & 0.788$\pm$0.007(6.0) & 0.904$\pm$0.002(7.0) & 0.981$\pm$0.001(8.0) & \textbf{0.994$\pm$0.001(1.5)} & 6.44 \\\hline
\multirow{5}{*}{FT\_T} & \textit{random} & 0.819$\pm$0.005(21.0) & 0.726$\pm$0.009(7.0) & 0.711$\pm$0.006(5.0) & 0.946$\pm$0.004(18.0) & 0.750$\pm$0.009(18.0) & 0.902$\pm$0.001(8.0) & 0.959$\pm$0.001(13.0) & 0.982$\pm$0.007(11.0) & 12.62 \\
 & \textit{uncertainty} & 0.847$\pm$0.012(16.0) & 0.725$\pm$0.007(8.0) & 0.706$\pm$0.011(7.0) & 0.978$\pm$0.004(3.0) & 0.775$\pm$0.022(10.0) & 0.908$\pm$0.004(5.0) & 0.985$\pm$0.003(2.0) & 0.993$\pm$0.004(3.0) & 6.75 \\
 & \textit{ID} & 0.844$\pm$0.009(17.0) & 0.721$\pm$0.007(11.0) & 0.698$\pm$0.014(11.0) & 0.975$\pm$0.003(6.0) & 0.772$\pm$0.017(11.0) & 0.905$\pm$0.004(6.0) & \textbf{0.987$\pm$0.001(1.0)} & 0.992$\pm$0.005(4.0) & 8.38 \\
 & \textit{EGAL} & 0.805$\pm$0.005(23.0) & 0.656$\pm$0.038(22.0) & 0.666$\pm$0.007(21.0) & 0.942$\pm$0.004(23.0) & 0.740$\pm$0.013(23.0) & 0.898$\pm$0.005(11.0) & 0.963$\pm$0.003(10.0) & 0.985$\pm$0.004(10.0) & 17.88 \\
 & \textit{QBC} & 0.851$\pm$0.005(15.0) & 0.730$\pm$0.008(5.0) & 0.702$\pm$0.014(9.0) & 0.975$\pm$0.001(8.5) & 0.793$\pm$0.005(4.0) & 0.909$\pm$0.003(4.0) & 0.985$\pm$0.000(5.0) & \textbf{0.994$\pm$0.001(1.5)} & 6.50 \\\hline
\multirow{5}{*}{LDA} & \textit{random} & 0.772$\pm$0.006(28.0) & 0.611$\pm$0.006(26.0) & 0.669$\pm$0.005(20.0) & 0.884$\pm$0.012(29.0) & 0.733$\pm$0.006(27.0) & 0.680$\pm$0.003(27.0) & 0.854$\pm$0.004(29.0) & 0.858$\pm$0.006(29.0) & 26.88 \\
 & \textit{uncertainty} & 0.791$\pm$0.006(27.0) & 0.613$\pm$0.010(25.0) & 0.673$\pm$0.015(18.0) & 0.922$\pm$0.012(26.0) & 0.754$\pm$0.011(16.0) & 0.671$\pm$0.013(28.0) & 0.877$\pm$0.017(26.0) & 0.881$\pm$0.011(27.0) & 24.12 \\
 & \textit{ID} & 0.796$\pm$0.005(26.0) & 0.606$\pm$0.006(28.0) & 0.672$\pm$0.005(19.0) & 0.914$\pm$0.010(28.0) & 0.750$\pm$0.010(17.0) & 0.620$\pm$0.008(29.0) & 0.862$\pm$0.010(28.0) & 0.863$\pm$0.012(28.0) & 25.38 \\
 & \textit{EGAL} & 0.760$\pm$0.007(30.0) & 0.600$\pm$0.008(29.0) & 0.648$\pm$0.009(28.0) & 0.858$\pm$0.014(30.0) & 0.720$\pm$0.008(29.0) & 0.611$\pm$0.015(30.0) & 0.835$\pm$0.006(30.0) & 0.805$\pm$0.018(30.0) & 29.50 \\
 & \textit{QBC} & 0.770$\pm$0.009(29.0) & 0.607$\pm$0.009(27.0) & 0.675$\pm$0.008(16.0) & 0.917$\pm$0.008(27.0) & 0.763$\pm$0.008(14.0) & 0.689$\pm$0.006(26.0) & 0.901$\pm$0.005(22.0) & 0.905$\pm$0.006(24.0) & 23.12 \\\hline
\multirow{5}{*}{TF-IDF} & \textit{random} & 0.837$\pm$0.003(18.0) & 0.708$\pm$0.012(16.0) & 0.666$\pm$0.006(22.0) & 0.945$\pm$0.002(20.0) & 0.740$\pm$0.011(24.5) & 0.808$\pm$0.007(18.0) & 0.887$\pm$0.011(24.0) & 0.912$\pm$0.014(23.0) & 20.69 \\
 & \textit{uncertainty} & 0.871$\pm$0.005(3.0) & 0.719$\pm$0.009(13.0) & 0.684$\pm$0.006(13.0) & 0.975$\pm$0.001(8.5) & 0.758$\pm$0.017(15.0) & 0.807$\pm$0.016(19.0) & 0.919$\pm$0.017(18.0) & 0.943$\pm$0.021(18.0) & 13.44 \\
 & \textit{ID} & 0.862$\pm$0.004(6.0) & 0.696$\pm$0.004(17.0) & 0.683$\pm$0.003(14.0) & 0.971$\pm$0.002(11.5) & 0.744$\pm$0.013(22.0) & 0.803$\pm$0.016(20.0) & 0.915$\pm$0.010(19.0) & 0.926$\pm$0.018(20.5) & 16.25 \\
 & \textit{EGAL} & 0.800$\pm$0.008(24.0) & 0.642$\pm$0.043(24.0) & 0.637$\pm$0.009(29.0) & 0.942$\pm$0.005(22.0) & 0.745$\pm$0.009(21.0) & 0.809$\pm$0.008(17.0) & 0.895$\pm$0.012(23.0) & 0.912$\pm$0.016(22.0) & 22.75 \\
 & \textit{QBC} & 0.854$\pm$0.012(10.0) & 0.713$\pm$0.007(15.0) & 0.676$\pm$0.007(15.0) & 0.965$\pm$0.002(14.0) & 0.778$\pm$0.005(7.0) & 0.812$\pm$0.005(16.0) & 0.932$\pm$0.004(16.0) & 0.952$\pm$0.007(16.0) & 13.62 \\\hline
\multirow{5}{*}{TF} & \textit{random} & 0.832$\pm$0.004(19.0) & 0.674$\pm$0.009(21.0) & 0.651$\pm$0.007(27.0) & 0.943$\pm$0.003(21.0) & 0.729$\pm$0.007(28.0) & 0.802$\pm$0.007(22.0) & 0.880$\pm$0.012(25.0) & 0.902$\pm$0.017(25.0) & 23.50 \\
 & \textit{uncertainty} & 0.868$\pm$0.003(4.0) & 0.683$\pm$0.008(19.0) & 0.660$\pm$0.007(24.0) & 0.973$\pm$0.001(10.0) & 0.746$\pm$0.016(20.0) & 0.797$\pm$0.022(23.0) & 0.911$\pm$0.012(20.0) & 0.937$\pm$0.029(19.0) & 17.38 \\
 & \textit{ID} & 0.867$\pm$0.002(5.0) & 0.694$\pm$0.004(18.0) & 0.675$\pm$0.002(17.0) & 0.970$\pm$0.001(13.0) & 0.748$\pm$0.011(19.0) & 0.796$\pm$0.008(24.0) & 0.911$\pm$0.006(21.0) & 0.926$\pm$0.018(20.5) & 17.19 \\
 & \textit{EGAL} & 0.799$\pm$0.016(25.0) & 0.559$\pm$0.011(30.0) & 0.604$\pm$0.010(30.0) & 0.937$\pm$0.002(25.0) & 0.716$\pm$0.026(30.0) & 0.795$\pm$0.015(25.0) & 0.863$\pm$0.011(27.0) & 0.900$\pm$0.017(26.0) & 27.25 \\
 & \textit{QBC} & 0.853$\pm$0.004(13.0) & 0.678$\pm$0.011(20.0) & 0.658$\pm$0.008(25.0) & 0.963$\pm$0.002(15.0) & 0.770$\pm$0.006(12.0) & 0.803$\pm$0.007(21.0) & 0.929$\pm$0.004(17.0) & 0.948$\pm$0.009(17.0) & 17.50
 \\\hline\hline
\end{tabular}%

}

\end{table}
\end{landscape}

\section{Results and Discussion}\label{sec:results}


To illustrate the performance differences observed between the different representations explored, Figures \ref{fig:MDCR} and \ref{fig:MRS} show the learning curves for each different representation (separated by selection strategy) for the MDCR and MRS datasets respectively.\footnote{Similar figures for the other 6 datasets can be found at \label{github}\emph{URL hidden} for anonymous review.} In these plots the horizontal axis denotes the number of instances labelled so far, and the vertical axis denotes the \emph{accuracy+} score achieved. It should be noted that each curve starts with 10 rather than 0 along the horizontal axis, corresponding to the initial seed labelling described earlier. 

Generally speaking, we can observe that better performance is achieved when active learning is used in combination with a text representations based on word embeddings rather than the simpler vector-based text representations (i.e. TF and TF-IDF) and those based on topic modelling (i.e. LDA). More specifically, in Figure \ref{fig:MDCR}, we observe that BERT consistently outperforms any other representation by a reasonably large margin across all selection strategies. Another interesting observation is that FastText, FastText\_trained and TF-IDF have similar performance, and LDA performs worst across all situations. In Figure \ref{fig:MRS}, we see a similar pattern that, in the majority of cases, the performance of the approaches based on BERT surpass the performances achieved using other representations (except for EGAL where FastText\_trained gives the best performance). Besides, the remaining two word embeddings (i.e. FastText and FastText\_trained) behave close to BERT in many selection strategies, exceeding TF and TF-IDF by a large margin. Again, LDA performs poorly when used in combination with all selection strategies.

We summarize the results of all methods in Table \ref{tab:AUC_summary}. In this table, each column denotes the performance of different active learning processes on a specific dataset. Different representation and selection strategy combinations are compared and the best results achieved for each dataset are highlighted. The numbers in bracket stands for the ranking of each method when compared to the performance of the other approaches for a specific dataset and the last column reports the average ranking of each representation-selection-strategy combination, where a smaller number means a higher rank. 

Table \ref{tab:AUC_summary} presents a very clear message that the word embedding representations perform well across all datasets, which is evidenced by the relatively higher ranks as compared to TF, TF-IDF and LDA. Overall, BERT is the best performing representation with average ranks of 2.81 for BERT + uncertainty, 4.5 for BERT + information density, and 4.81 for BERT + QBC being the highest average ranks overall. 

As suggested by \cite{benavoli2016should}, Wilcoxon signed-rank tests have been performed for pairwise comparisons of the mean ranks between methods. The full results of these tests are too large to include in this paper, but are available in online supplementary material \footnote{\emph{URL hidden} for anonymous review.}. As QBC is the best performing selection strategy overall, we include the results significance tests comparing the performance of the different representation methods when the QBC selection strategy is used across all datasets in Table \ref{tab:p_values_qbc}. This shows the win/loss/tie count for each pair of representations and the p-value from the related significance test. The table demonstrate that all embedding-based methods are significantly different from methods based on TF, TF-IDF and LDA with $p<0.05$. However, embedding-based methods do not have significant difference between each other. Remarkably, BERT achieves the most wins as compared to any other representations.

\begin{table}[!htbp]
\centering
\caption{P values and win/draw/lose of pairwise comparison of QBC-based methods.}
\label{tab:p_values_qbc}
\begin{tabular}{p{15mm} p{15mm} p{15mm} p{15mm} p{15mm} p{15mm} p{15mm}}\hline\hline
            & BERT       & FT         & FT\_T       & LDA        & TF-IDF     & TF \\\hline
BERT   &                 & 6/0/2           & 5/0/3           & 8/0/0           & 8/0/0           & 8/0/0   \\
FT     & 0.0687          &                 & 3/1/4           & 8/0/0           & 8/0/0           & 8/0/0   \\
FT\_T   & 0.0929          & 0.4982          &                 & 8/0/0           & 7/0/1           & 7/0/1   \\
LDA    & {\fontseries{b}\selectfont 0.0117} & {\fontseries{b}\selectfont0.0117} & {\fontseries{b}\selectfont0.0117} &                 & 0/0/8           & 1/0/7   \\
TF-IDF & {\fontseries{b}\selectfont0.0117} & {\fontseries{b}\selectfont0.0117} & {\fontseries{b}\selectfont0.0173} & {\fontseries{b}\selectfont0.0117} &                 & 7/0/1   \\
TF     & {\fontseries{b}\selectfont0.0117} & {\fontseries{b}\selectfont0.0116} & {\fontseries{b}\selectfont0.0173} & {\fontseries{b}\selectfont0.0251} & {\fontseries{b}\selectfont0.0117} &\\ \hline\hline       
\end{tabular}

\end{table}

\begin{figure*}[!htbp]
\centering
  \subfigure[MDCR: BERT \label{fig:tsne_bert_MDCR}]{\includegraphics[width=0.3\linewidth]{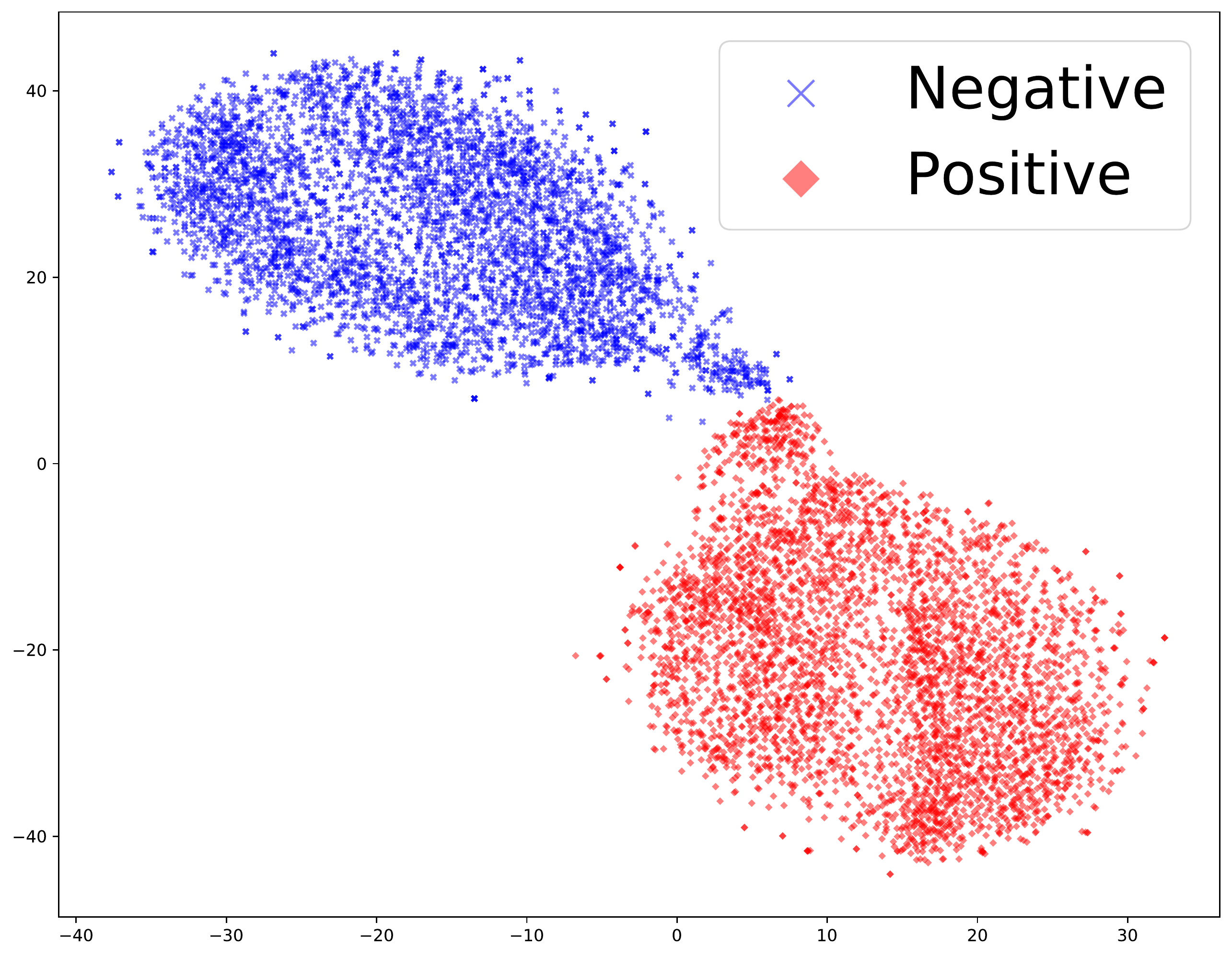}}
  \subfigure[MDCR: TF-IDF \label{fig:tsne_tfidf_MDCR}]{\includegraphics[width=0.3\linewidth]{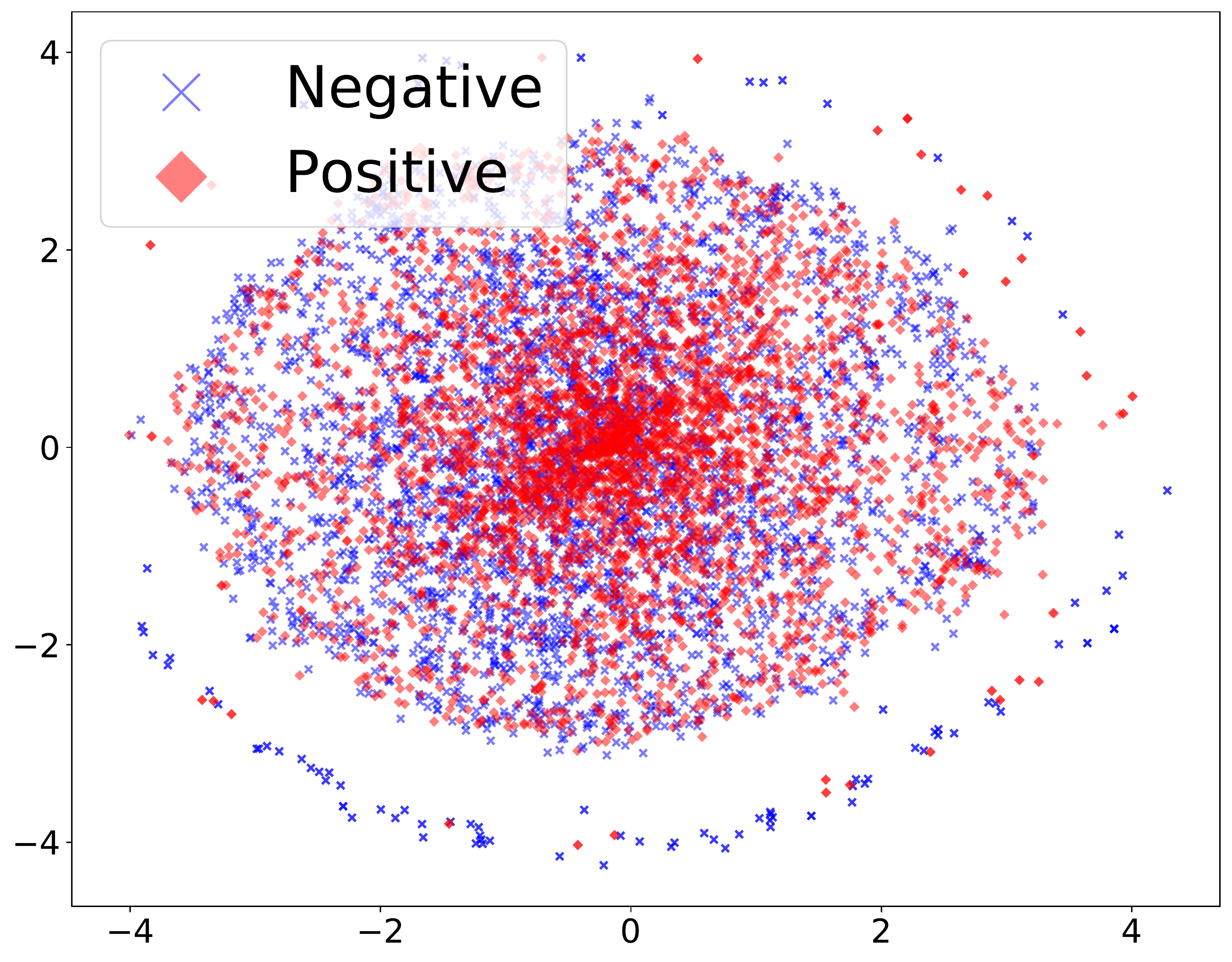}}
   \subfigure[MDCR: LDA \label{fig:tsne_lda_MDCR}]{\includegraphics[width=0.3\linewidth]{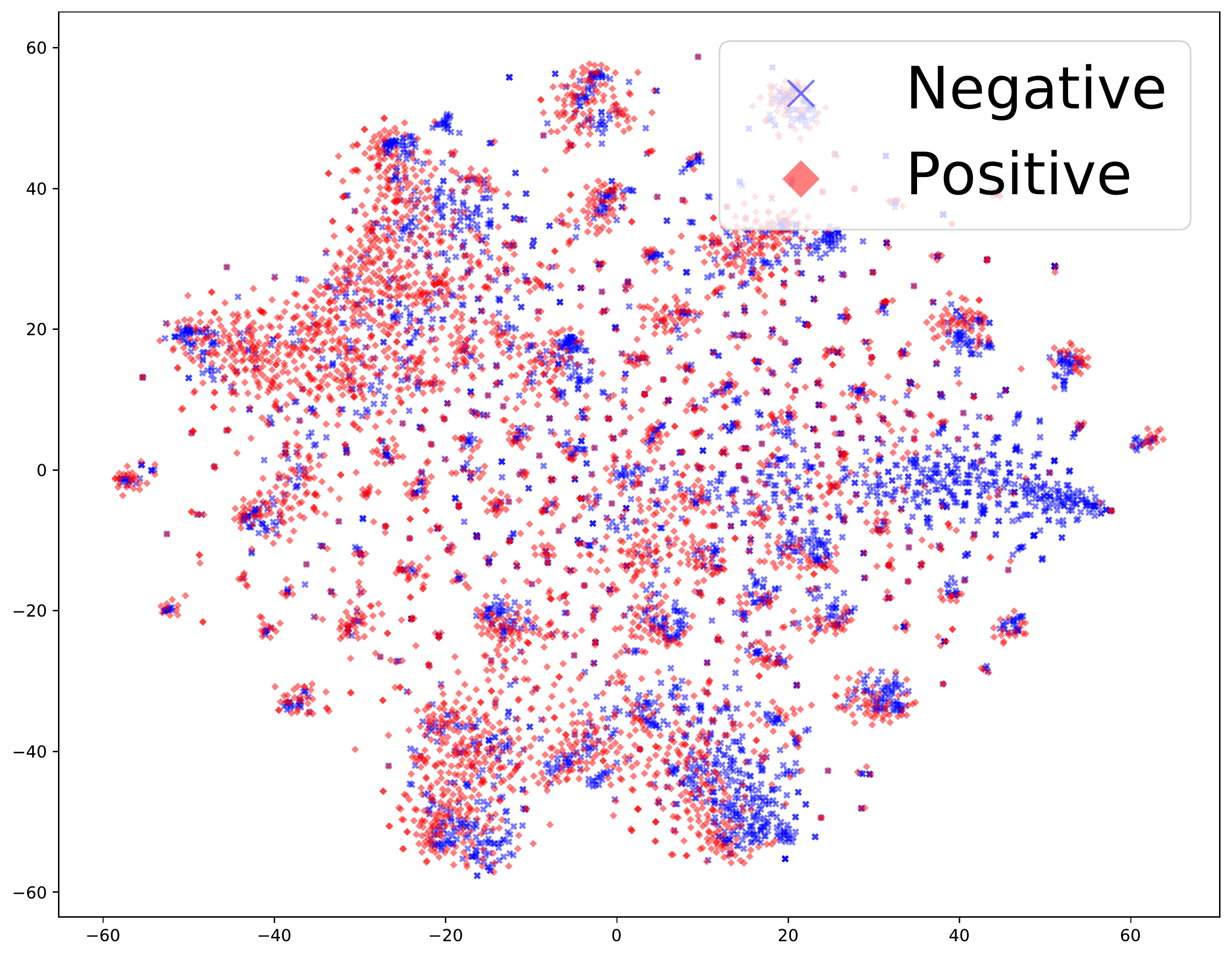}}
 \centering
  \subfigure[MRS: BERT \label{fig:tsne_bert_MRS}]{\includegraphics[width=0.3\linewidth]{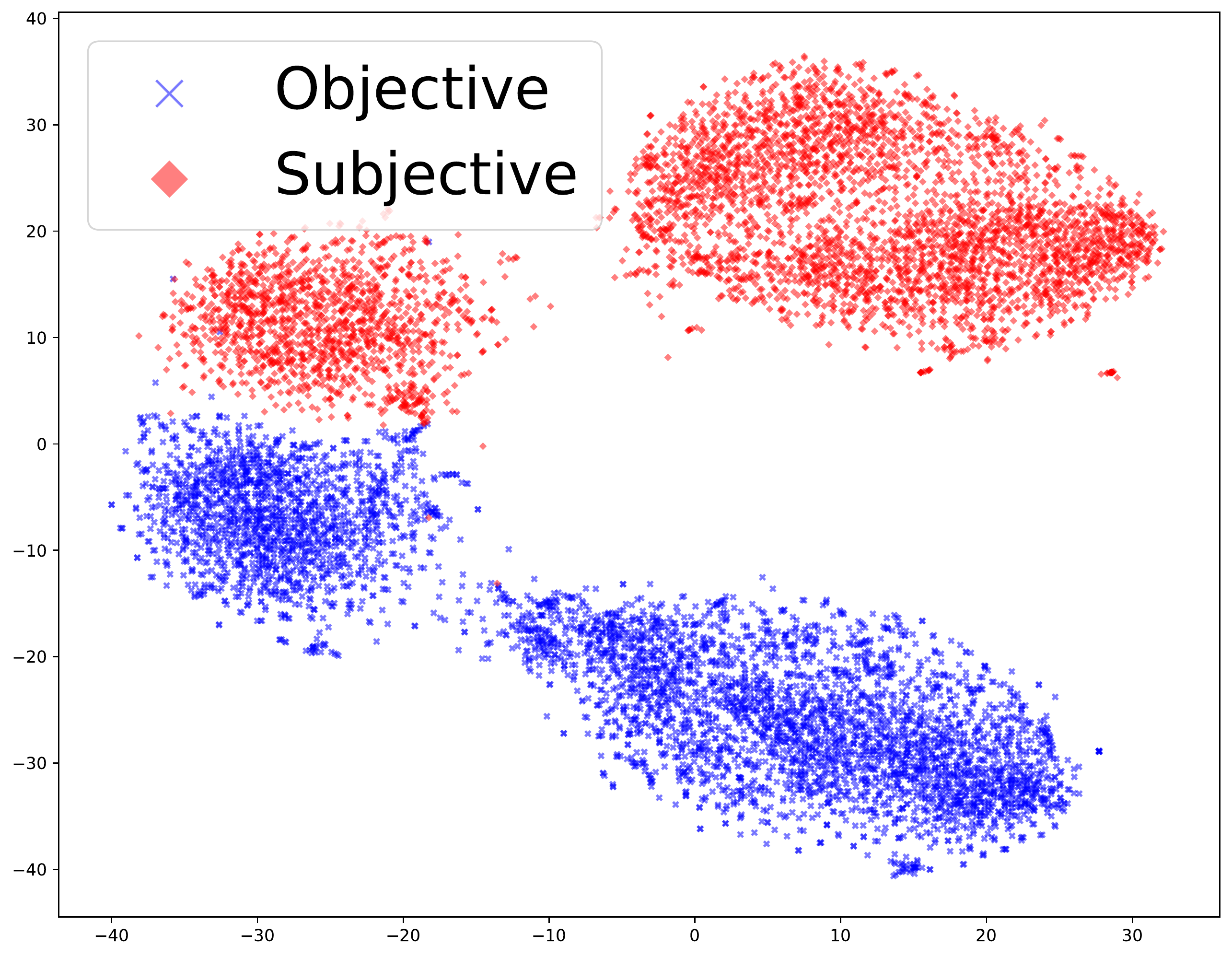}}
  \subfigure[MRS: TF-IDF \label{fig:tsne_tfidf_MRS}]{\includegraphics[width=0.3\linewidth]{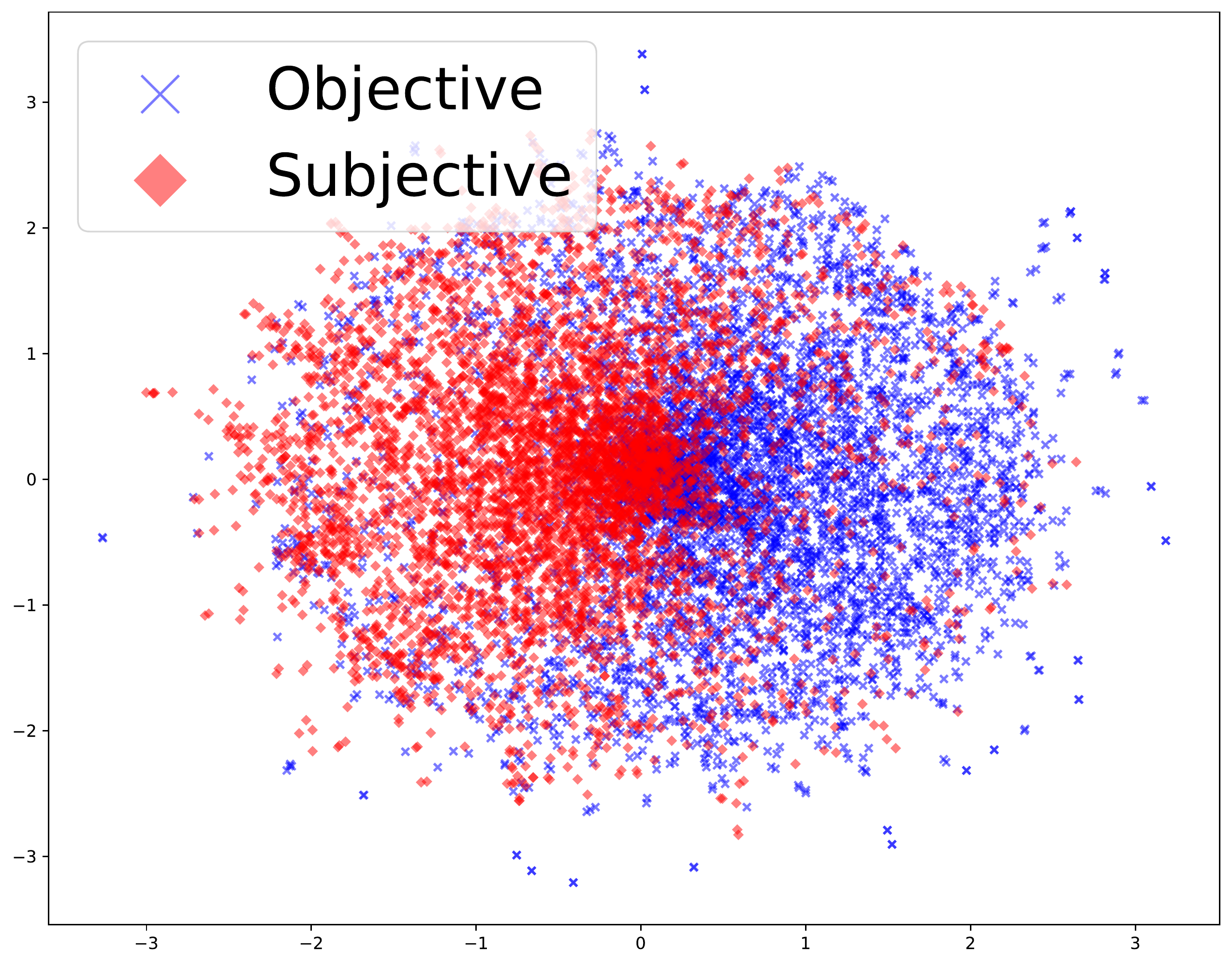}}
   \subfigure[MRS: LDA \label{fig:tsne_lda_MRS}]{\includegraphics[width=0.3\linewidth]{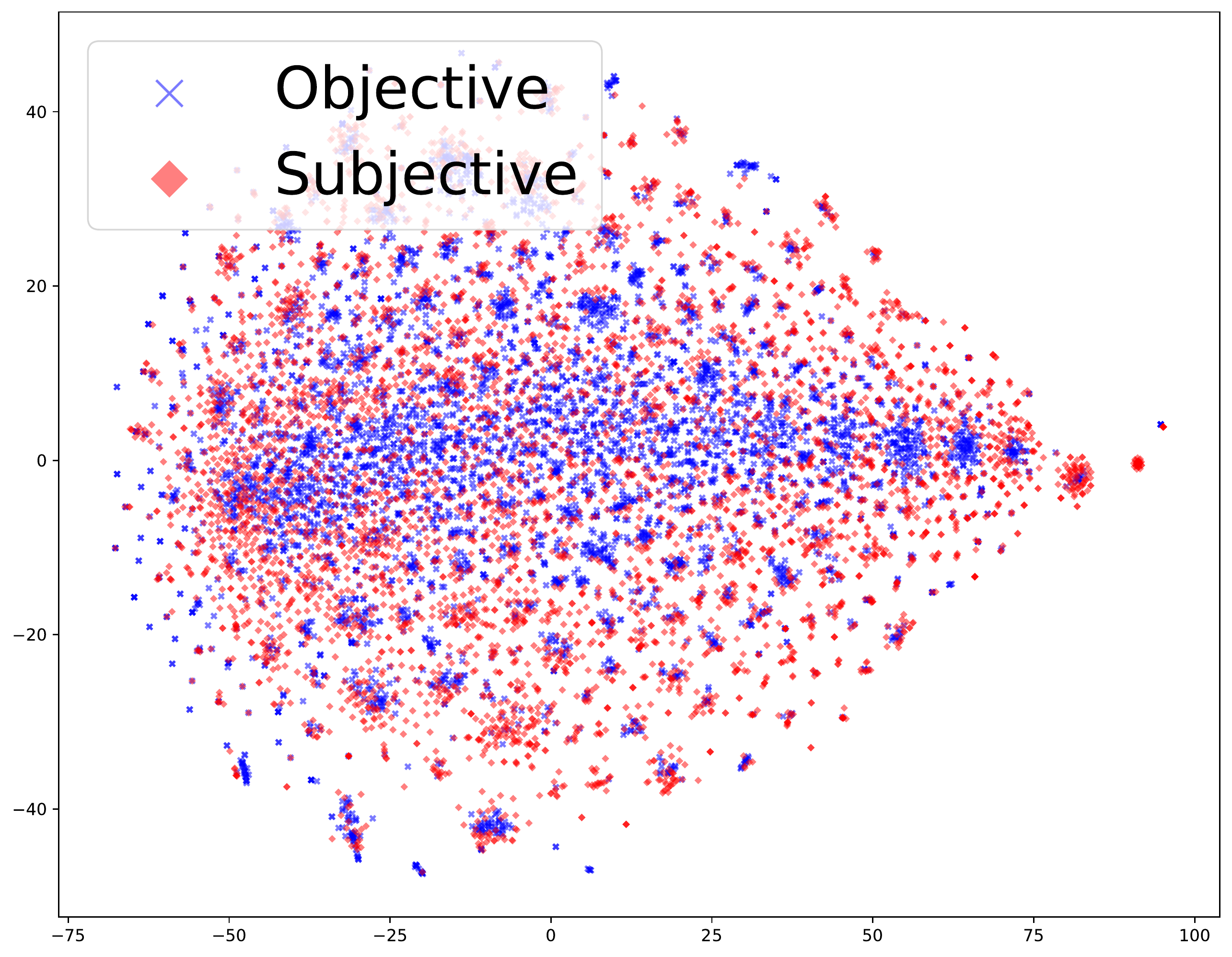}}
\medskip
  \caption[]{T-SNE visualisations of movie reviews and customer reviews regarding \emph{(MDCR)} (Figure \ref{fig:tsne_bert_MDCR}, \ref{fig:tsne_tfidf_MDCR}, \ref{fig:tsne_lda_MDCR}) and \emph{(MRS)} (Figure \ref{fig:tsne_bert_MRS}, \ref{fig:tsne_tfidf_MRS}, \ref{fig:tsne_lda_MRS}) dataset in the corresponding feature space respectively. Red squares and blue crosses indicate reviews of different classes.}
  \label{fig:tsne_rep}
  
\end{figure*}



\subsection{Analysis of Different Representations}

The previous experiments have illustrated the superior performance of word embeddings, especially BERT, in active learning. To provide some insight into the impacts of using different representations, Figure \ref{fig:tsne_rep} shows visualisations of instances from the Multi Domain Customer Review (MDCR) and Movie Review Subjectivity (MRS) datasets generated using t-SNE \cite{maaten2008visualizing} based BERT, TF-IDF and LDA representations. Instances are coloured according to the class to which they belong. We can see that for the BERT representation instances of the same class tend to cluster near to each other and that there is good separation between instances of the two classes (see Figure \ref{fig:tsne_bert_MDCR} and \ref{fig:tsne_bert_MRS}), even though no label information is used in generating the BERT representation or these visualisations.  For the equivalent  TF-IDF (Figures \ref{fig:tsne_tfidf_MDCR} and \ref{fig:tsne_tfidf_MRS}) and LDA (Figures \ref{fig:tsne_lda_MDCR} and \ref{fig:tsne_lda_MRS}), visualisations classes are less well clustered and overlap much more. This ability of BERT to generate instances representations that are easily separable is likely to contribute strongly towards its ability to lead to highly performing active learning systems. We suppose that the unsatisfactory performance of the LDA-based representation indicates each class is likely to contain a mixture of most of the topics in these datasets.


\section{Conclusions}\label{sec:conclusions}

Active learning processes used with text data rely heavily on the document representation mechanism used. This paper presented an evaluation experiment which explored the effectiveness of different text representations in an active learning context. The performance of different text representation techniques combined with popular selection strategies was compared over datasets from different domains to investigate a general active learning framework for data labelling tasks. The comparison showed that the embedding-based representations, which are rarely used in active learning, lead to better performance compared to vector based representations. Several of the most commonly used selection strategies have been applied in experiments to mitigate the impact of specific selection strategies on the effectiveness of different text representations. Notably, BERT combined with uncertainty sampling greatly facilitates the application of active learning for text labelling. Hence, we suggest that BERT with uncertainty sampling is the default framework while BERT with QBC/ID and FastText\_trained with QBC can be alternatives for text classification in the context of labelling task in some cases.

An important application of active learning is to labelling the included/ex-cluded studies in literature review \cite{wallace2010active,hashimoto2016topic,miwa2014reducing} which is usually an imbalanced dataset. So It leads to more exploration of the active learning framework over an imbalanced dataset in future work.

\clearpage
%
%
%

\begin{thebibliography}{8}
\bibitem{settles2009active}
Settles, B., 2009. Active learning literature survey. University of Wisconsin-Madison Department of Computer Sciences.

\bibitem{tong2001support}
Tong, S. and Chang, E., 2001, October. Support vector machine active learning for image retrieval. In Proceedings of the ninth ACM international conference on Multimedia (pp. 107-118). ACM.

\bibitem{zhang2002active}
Zhang, C. and Chen, T., 2002. An active learning framework for content-based information retrieval. IEEE transactions on multimedia, 4(2), pp.260-268.

\bibitem{hoi2006large}
Hoi, S.C., Jin, R. and Lyu, M.R., 2006, May. Large-scale text categorization by batch mode active learning. In Proceedings of the 15th international conference on World Wide Web (pp. 633-642). ACM.

\bibitem{Liere1997active}
Liere, R. and Tadepalli, P., 1997, July. Active learning with committees for text categorization. In AAAI/IAAI (pp. 591-596).

\bibitem{zhang2017active}
Zhang, Y., Lease, M. and Wallace, B.C., 2017, February. Active discriminative text representation learning. In Thirty-First AAAI Conference on Artificial Intelligence.

\bibitem{zhao2017deep}
Zhao, W., 2017. Deep Active Learning for Short-Text Classification.

\bibitem{singh2018improving}
Singh, G., Thomas, J. and Shawe-Taylor, J., 2018. Improving active learning in systematic reviews. arXiv preprint arXiv:1801.09496.


\bibitem{tur2005combining}
Tur, G., Hakkani-Tür, D. and Schapire, R.E., 2005. Combining active and semi-supervised learning for spoken language understanding. Speech Communication, 45(2), pp.171-186.


\bibitem{hu2010egal}
Hu, R., Delany, S.J. and Mac Namee, B., 2010, July. EGAL: Exploration guided active learning for TCBR. In International Conference on Case-Based Reasoning (pp. 156-170). Springer, Berlin, Heidelberg.

\bibitem{hu2008sweetening}
Hu, R., Mac Namee, B. and Delany, S.J., 2008. Sweetening the dataset: Using active learning to label unlabelled datasets.

\bibitem{wallace2010active}
Wallace, B.C., Small, K., Brodley, C.E. and Trikalinos, T.A., 2010, July. Active learning for biomedical citation screening. In Proceedings of the 16th ACM SIGKDD international conference on Knowledge discovery and data mining (pp. 173-182). ACM.

\bibitem{mikolov2013efficient}
Mikolov, T., Chen, K., Corrado, G. and Dean, J., 2013. Efficient estimation of word representations in vector space. arXiv preprint arXiv:1301.3781.

\bibitem{pennington2014glove}
Pennington, J., Socher, R. and Manning, C., 2014, October. Glove: Global vectors for word representation. In Proceedings of the 2014 conference on empirical methods in natural language processing (EMNLP) (pp. 1532-1543).

\bibitem{bojanowski2017enriching}
Bojanowski, P., Grave, E., Joulin, A. and Mikolov, T., 2017. Enriching word vectors with subword information. Transactions of the Association for Computational Linguistics, 5, pp.135-146.

\bibitem{joulin2016bag}
Joulin, A., Grave, E., Bojanowski, P. and Mikolov, T., 2016. Bag of tricks for efficient text classification. arXiv preprint arXiv:1607.01759.

\bibitem{joulin2016fasttext}
Joulin, A., Grave, E., Bojanowski, P., Douze, M., Jégou, H. and Mikolov, T., 2016. Fasttext. zip: Compressing text classification models. arXiv preprint arXiv:1612.03651.

\bibitem{howard2018universal}
Howard, J. and Ruder, S., 2018. Universal language model fine-tuning for text classification. arXiv preprint arXiv:1801.06146.

\bibitem{radford2018improving}
Radford, A., Narasimhan, K., Salimans, T. and Sutskever, I., 2018. Improving language understanding by generative pre-training. URL https://s3-us-west-2. amazonaws. com/openai-assets/researchcovers/languageunsupervised/language understanding paper. pdf.

\bibitem{devlin2018bert}
Devlin, J., Chang, M.W., Lee, K. and Toutanova, K., 2018. Bert: Pre-training of deep bidirectional transformers for language understanding. arXiv preprint arXiv:1810.04805.

\bibitem{hashimoto2016topic}
Hashimoto, K., Kontonatsios, G., Miwa, M. and Ananiadou, S., 2016. Topic detection using paragraph vectors to support active learning in systematic reviews. Journal of biomedical informatics, 62, pp.59-65.

\bibitem{siddhant2018deep}
Siddhant, A. and Lipton, Z.C., 2018. Deep Bayesian active learning for natural language processing: Results of a large-scale empirical study. arXiv preprint arXiv:1808.05697.

\bibitem{miwa2014reducing}
Miwa, M., Thomas, J., O’Mara-Eves, A. and Ananiadou, S., 2014. Reducing systematic review workload through certainty-based screening. Journal of biomedical informatics, 51, pp.242-253.

\bibitem{mo2015supporting}
Mo, Y., Kontonatsios, G. and Ananiadou, S., 2015. Supporting systematic reviews using LDA-based document representations. Systematic reviews, 4(1), p.172.

\bibitem{zhu2008active}
Zhu, Jingbo, Huizhen Wang, Tianshun Yao, and Benjamin K. Tsou. "Active learning with sampling by uncertainty and density for word sense disambiguation and text classification." In Proceedings of the 22nd International Conference on Computational Linguistics-Volume 1, pp. 1137-1144. Association for Computational Linguistics, 2008.

\bibitem{lewis1994sequential}
Lewis, D.D. and Gale, W.A., 1994. A sequential algorithm for training text classifiers. In SIGIR’94 (pp. 3-12). Springer, London.

\bibitem{blei2003latent}
Blei, D.M., Ng, A.Y. and Jordan, M.I., 2003. Latent dirichlet allocation. Journal of machine Learning research, 3(Jan), pp.993-1022.

\bibitem{shannon1948mathematical}
Shannon, C.E., 1948. A mathematical theory of communication. Bell system technical journal, 27(3), pp.379-423.

\bibitem{settles2008analysis}
Settles, B. and Craven, M., 2008, October. An analysis of active learning strategies for sequence labeling tasks. In Proceedings of the conference on empirical methods in natural language processing (pp. 1070-1079). Association for Computational Linguistics.

\bibitem{seung1992query}
Seung, H.S., Opper, M. and Sompolinsky, H., 1992, July. Query by committee. In Proceedings of the fifth annual workshop on Computational learning theory (pp. 287-294). ACM.

\bibitem{dagan1995committee}
Dagan, I. and Engelson, S.P., 1995. Committee-based sampling for training probabilistic classifiers. In Machine Learning Proceedings 1995 (pp. 150-157). Morgan Kaufmann.

\bibitem{hsu2003practical}
Hsu, C.W., Chang, C.C. and Lin, C.J., 2003. A practical guide to support vector classification.

\bibitem{tong2000support}
Tong, S. and Koller, D., 2000. Support vector machine active learning with applications to text classification. Journal of machine learning research, 2(Nov), pp.45-66.

\bibitem{mamitsuka1998query}
Mamitsuka, N.A.H., 1998, July. Query learning strategies using boosting and bagging. In Machine learning: proceedings of the fifteenth international conference (ICML’98) (Vol. 1). Morgan Kaufmann Pub.

\bibitem{pang2004sentimental}
Pang, B. and Lee, L., 2004, July. A sentimental education: Sentiment analysis using subjectivity summarization based on minimum cuts. In Proceedings of the 42nd annual meeting on Association for Computational Linguistics (p. 271). Association for Computational Linguistics.

\bibitem{blitzer2007biographies}
Blitzer, J., Dredze, M. and Pereira, F., 2007, June. Biographies, bollywood, boom-boxes and blenders: Domain adaptation for sentiment classification. In Proceedings of the 45th annual meeting of the association of computational linguistics (pp. 440-447).

\bibitem{belford2018stability}
Belford, M., Mac Namee, B. and Greene, D., 2018. Stability of topic modeling via matrix factorization. Expert Systems with Applications, 91, pp.159-169.

\bibitem{ding2008holistic}
Ding, X., Liu, B. and Yu, P.S., 2008, February. A holistic lexicon-based approach to opinion mining. In Proceedings of the 2008 international conference on web search and data mining (pp. 231-240). ACM.

\bibitem{mukherjee2010improving}
Mukherjee, A. and Liu, B., 2010, October. Improving gender classification of blog authors. In Proceedings of the 2010 conference on Empirical Methods in natural Language Processing (pp. 207-217). Association for Computational Linguistics.

\bibitem{zhang2015character}
Zhang, X., Zhao, J. and LeCun, Y., 2015. Character-level convolutional networks for text classification. In Advances in neural information processing systems (pp. 649-657).

\bibitem{benavoli2016should}
Benavoli, A., Corani, G. and Mangili, F., 2016. Should we really use post-hoc tests based on mean-ranks?. The Journal of Machine Learning Research, 17(1), pp.152-161.

\bibitem{maaten2008visualizing}
Maaten, L.V.D. and Hinton, G., 2008. Visualizing data using t-SNE. Journal of machine learning research, 9(Nov), pp.2579-2605.

\bibitem{mccann2017learned}
McCann, B., Bradbury, J., Xiong, C. and Socher, R., 2017. Learned in translation: Contextualized word vectors. In Advances in Neural Information Processing Systems (pp. 6294-6305).

\bibitem{peters2018deep}
Peters, M.E., Neumann, M., Iyyer, M., Gardner, M., Clark, C., Lee, K. and Zettlemoyer, L., 2018. Deep contextualized word representations. arXiv preprint arXiv:1802.05365.

\bibitem{zhang2019neural}
Zhang, Y., 2019. Neural NLP models under low-supervision scenarios (Doctoral dissertation).

\bibitem{lecun2015deep}
LeCun, Y., Bengio, Y. and Hinton, G., 2015. Deep learning. nature, 521(7553), p.436.

\bibitem{jones2004statistical}
Jones, K.S., 2004. A statistical interpretation of term specificity and its application in retrieval. Journal of documentation.

\bibitem{vaswani2017attention}
Vaswani, A., Shazeer, N., Parmar, N., Uszkoreit, J., Jones, L., Gomez, A.N., Kaiser, Ł. and Polosukhin, I., 2017. Attention is all you need. In Advances in neural information processing systems (pp. 5998-6008).

\bibitem{le2014distributed}
Le, Q. and Mikolov, T., 2014, January. Distributed representations of sentences and documents. In International conference on machine learning (pp. 1188-1196).


\end{thebibliography}
%

\clearpage

\end{document}